\documentclass{article}

\usepackage{microtype}
\usepackage{graphicx}
\usepackage{subcaption}
\usepackage{booktabs} % for professional tables
\usepackage{url}
\usepackage{xcolor}
\usepackage{soul}
\usepackage[table]{xcolor}
\usepackage[utf8]{inputenc} % allow utf-8 input
\usepackage[T1]{fontenc}    % use 8-bit T1 fonts
\usepackage{url}            % simple URL typesetting
\usepackage{booktabs}       % professional-quality tables
\usepackage{amsfonts}       % blackboard math symbols
\usepackage{nicefrac}       % compact symbols for 1/2, etc.
\usepackage{microtype}      % microtypography
\usepackage{xcolor}         % colors
\usepackage{amsmath}
\usepackage{graphicx}
\usepackage{enumitem}
\usepackage{tabularx}
\usepackage{multirow} 
\usepackage{adjustbox}
\usepackage{booktabs,adjustbox,array,makecell} % in the preamble
\usepackage{amsmath}
\usepackage{amssymb}
\usepackage{mathtools}
\usepackage{amsthm}
\usepackage{threeparttable}

\usepackage{hyperref}

\usepackage[accepted]{icml2026}

\usepackage{amsmath}
\usepackage{amssymb}
\usepackage{mathtools}
\usepackage{amsthm}

\usepackage[capitalize,noabbrev]{cleveref}

\theoremstyle{plain}

\theoremstyle{definition}

\theoremstyle{remark}

\usepackage[textsize=tiny]{todonotes}

\icmltitlerunning{Beyond Rewards in RL for Cyber Defence}

\begin{document}

\twocolumn[
  \icmltitle{Beyond Rewards in RL for Cyber Defence}



\icmlsetsymbol{equal}{*}

\begin{icmlauthorlist}
\icmlauthor{Elizabeth Bates}{equal,turing}
\icmlauthor{Chris Hicks}{equal,turing}
\icmlauthor{Vasilios Mavroudis}{turing}
\end{icmlauthorlist}

\icmlaffiliation{turing}{The Alan Turing Institute, London, United Kingdom}

\icmlcorrespondingauthor{Elizabeth Bates}{ebates@turing.ac.uk}
\icmlcorrespondingauthor{Chris Hicks}{c.hicks@turing.ac.uk}
\icmlcorrespondingauthor{Vasilios Mavroudis}{vmavroudis@turing.ac.uk}

  \icmlkeywords{Machine Learning, ICML}

  \vskip 0.3in
]



\printAffiliationsAndNotice{\icmlEqualContribution}

\begin{abstract}
Recent years have seen an explosion of interest in autonomous cyber defence agents trained to defend computer networks using deep reinforcement learning. These agents are typically trained in cyber gym environments using dense, highly engineered reward functions which combine many penalties and incentives for a range of (un)desirable states and costly actions. Dense rewards help alleviate the challenge of exploring complex environments but risk biasing agents towards suboptimal and potentially riskier solutions, a critical issue in complex cyber environments. We thoroughly evaluate the impact of reward function structure on learning and policy behavioural characteristics using a variety of sparse and dense reward functions, two well-established cyber gyms, a range of network sizes, and both policy gradient and value-based RL algorithms. Our evaluation is enabled by a novel ground truth evaluation approach which allows directly comparing between different reward functions, illuminating the nuanced inter-relationships between rewards, action space, and the risks of suboptimal policies in cyber environments. Our results show that sparse rewards, provided they are goal aligned, and can be encountered frequently, uniquely offer both enhanced training reliability and more effective cyber defence agents with lower-risk policies. 
Surprisingly, sparse rewards can also yield policies that are better aligned with cyber defender goals and make sparing use of costly defensive actions without explicit reward-based numerical penalties. 
\end{abstract}

\section{Introduction}
% Motivate ACD
Cyber attacks are increasingly frequent, sophisticated, and automated, straining limited cyber defence resources and threatening critical digital systems that people depend upon worldwide. There has been a rising level of interest in using machine learning (ML) methods to improve cyber security; in particular deep reinforcement learning (DRL) which has the ability to learn complex policies from interaction alone, enabling the discovery of strategies unconstrained by flawed system or security models. DRL based autonomous cyber defence (ACD) agents, which have gathered much attention in the literature, could discover improved strategies and provide automation for tasks that currently occupy human analysts.

% DRL is a promising candidate for this task because of its ability to learn complex policies directly from interactions with an environment~\citep{sutton2018rl}, allowing for the discovery of strategies that are unconstrained by explicit models of security which may be incomplete or mistaken. 

% Role of cyber gyms
Cyber gyms provide efficient and controlled environments for ACD agents. This is particularly important for network security tasks, enabling the large number of interactions required for training without risking production networks or systems. Accordingly, many cyber gyms have been created to enable training agents that defend networked systems~\citep{vyas2023automated}. Cyber gyms define one or more Markov Decision Processes (MDPs) in terms of a state space comprising network and host information, an action space of defensive activities, and a reward function aligned to defensive objectives. ACD reward functions are typically highly engineered based on human judgment, combining multiple penalties and incentives determined for a variety of defensive actions and network states~\citep{yawningtitan, cyborg_acd_2021}. Dense rewards may be preferable because of expedited learning, providing apparently effective solutions using fewer environment steps during training, but they also risk constraining agents to sub-optimal solutions~\citep{riedmiller18a}. This is especially concerning for ACD agents which might then contain avoidable weaknesses that are difficult to identify in advance of an attack. Furthermore, dense rewards draw potentially arbitrary numerical equivalences between network states and actions. As the scale and complexity of cyber tasks grow this becomes increasingly challenging to manage and the risks of undesirable agent behaviour are exacerbated.

At the expense of generally requiring more training iterations, sparse rewards place fewer constraints on the solution space and could enable preferable or more effective policies to be discovered. Existing work has not investigated the effect that dense rewards might have on the performance of ACD agents trained using DRL. To investigate this possibility, and summarise the main contributions of this work, we: (1) propose a ground truth scoring mechanism for network security cyber gyms which allows a direct comparison between agents trained using different reward functions, (2) evaluate a comprehensive range of sparse and dense reward functions using two popular cyber gyms which are adapted to illustrate our ground truth mechanism, and (3) show that sparse reward functions can enhance the effectiveness, reliability, and risk-profiles of ACD agents across a variety of network sizes and topologies, action spaces, MDP models, and DRL algorithms.

%TODO: Clarify contributions in a list. --

% that are challenging to weigh in practice

%\subsection{Evaluating above the Abstraction}

\section{Background}
Here we introduce ACD, motivate evaluating more comprehensively, and define the key metrics we later build upon to fully evaluate the impact of reward functions in ACD.

\subsection{Autonomous Cyber Defence}\label{sec:background_acd}
% Why is ACD important or interesting
ACD agents aim to actively mitigate attacks on computer networks using ML techniques rather than traditional rule-based approaches. By alleviating the bottlenecks of human response speed and information processing, ACD agents could provide a much needed counterbalance to the ever-increasing scale and sophistication of cyber threats. 
% Why RL for ACD
Reinforcement learning (RL), and particularly DRL given the enormity of data generated by computer networks, is promising as it allows learning defensive strategies from interaction alone without the need for explicit models of how networks, systems, and attackers behave. Such models must continually be updated as attackers evolve, frequently undermining the tools and techniques that derive security proofs or assurances from their correctness. By observing the network state and choosing defensive actions, DRL agents can learn novel and adaptive strategies for defending computer networks that do not depend on potentially incorrect or outdated assumptions.

% The importance of rewards and the need for more accurate evaluation
Since their learning is guided by maximising long-term rewards, ACD agents critically depend on the rewards provided throughout training. Furthermore, the exploration required for learning from trial-and-error demands a cyber gym allowing extensive experimentation (i.e., risk-taking) without jeopardising valuable production systems. Many cyber gyms have been created~\citep{vyas2023automated}, provided publicly~\citep{microsoft_cyberbattlesim, oesch_24, yawningtitan}, and even used for competitions seeking the best performing agents~\citep{cyborg_acd_2021, hicksCAGE3_23, foleyCAGEII22}.
Despite these promising developments, previous work on ACD is limited to evaluating performance using only mean episodic rewards, and variance of the same, over a number of fixed-policy rollouts. Unlike games (e.g., chess) which correspond relatively naturally to the MDP framework, defending a network of computer hosts does not. Real-world attackers are not confined to turn-based interactions, partial observability affects many aspects of the network, and there is never a state where the defender can be definitively crowned the winner.

% Last para..
Most cyber gyms, and prominent ACD competitions, have hand-crafted dense reward functions that are used to train and evaluate agents. Such rewards may misrepresent the true performance of agents and it is impractical for them to accurately represent human knowledge~\citep{hu20shapereward}, biasing models towards possibly lower-performance and higher-risk strategies. We identify and, to the best of our knowledge, are the first to address the need for evaluation methods that consider the ground truth of complex cyber environments, i.e., beyond the modelling constraints of the RL framework. Our ground truth scoring mechanism permits a direct and reproducible comparison between different reward strategies, enabling experiments that empirically quantify the performance and risk characteristics of reward functions in ACD environments.

\subsection{Reliability and Risk in RL} \label{risk}

The reliability and risks of RL agents are critical issues for cyber defence applications where inconsistent performance can be costly or dangerous. Training reliability metrics measure how consistently an RL algorithm performs across multiple training runs, and risk metrics quantify expectations of worst-case performance. 

\subsubsection*{Training reliability}
\label{sec:background_reliability}

To evaluate the impact of reward function on training reliability, we build upon the quantitative RL training reliability metrics proposed by~\citet{chan2019measuring} based on dispersion variability i.e., the width of the mean episodic rewards distribution. %\hl{We also use the standard 95\% confidence interval but with use of the t-statistic, since the sample size for each experiment is 25.}

\paragraph{Dispersion variability across time (DT)} measures the stability of RL training across time. Smooth monotonic policy improvements offer the lowest $DT$ scores, indicating high reliability during training and lowered computational costs.
$DT$ is measured by averaging the inter-quartile range (IQR) within a sliding window along each detrended training curve. Detrending ensures positive trends in policy improvement do not influence the metric and is calculated using differencing of the training curve, $y^\prime_t = y_t - y_{t-1}$, where $y_t$ is the mean episodic reward at evaluation step $t$. Where $I$ denotes the total number of runs, the average DT across multiple runs is calculated: $$\bar{\text{DT}} = \frac{\sum^{I}_{i=0}\text{DT}_{i}}{I}$$

    %% DR 
\paragraph{Dispersion variability across runs (DR)} measures the reproducibility of RL training across multiple runs. Low DR indicates high consistency between training runs, meaning fewer total training runs are required to discover the best performing agents. DR is measured by averaging IQR across multiple training runs at each evaluation step, ensuring the metric captures differences resulting from random initialisation and environment stochasticity. Let $\bar{R_i}$ denote the mean episodic reward over some window of training run $i$, and $\{\bar{R_1}, \bar{R_2}, \ldots, \bar{R_I}\}$ the set of all such $\bar{R_i}$ across $I$ total runs, then:
    $$\text{DR} = \text{IQR}\big(\{\bar{R_i}\}^I_{i=1}\big)$$

%\paragraph{Confidence Intervals (CI)} 

\vspace{-2.5ex}
\subsubsection*{Risk after training} 
In ACD we are particularly concerned about the worst-case scenarios for a given agent. We calculate this by considering the worst-case expected loss across multiple rollouts of each trained policy.

\paragraph{Conditional Value at Risk (CVaR)} quantifies the risk associated with worst-case scenarios, defined by some quantile $\alpha$, i.e., expected performance in the worst $\alpha$ fraction of cases~\citep{acerbi_02}. By focussing on extreme values in the tails of the distribution, CVaR complements IQR methods in which they are cut off to focus on dispersion between central quartiles.

\paragraph{Risk across Fixed-Policy Rollouts (RF)} is calculated by applying CVaR to the distribution of multiple fixed-policy evaluation rollouts. Where $X = \{\bar{R_1}, \bar{R_2}, \ldots, \bar{R_I}\}$ denotes the set of mean episodic returns from the trained policy, and $\text{VaR}_\alpha$ the $\alpha$ quantile of $X$, then: $$\text{RF}_{\alpha}(X) = \text{CVaR}_{\alpha}(X) = \mathbb{E} [X \mid X \leq \text{VaR}_{\alpha}(X)]$$
%(i.e, the worst $\alpha$ fraction of episodic performances)

\section{Methodology}
Here we outline the methodology and experimental setup used to evaluate how different reward functions impact agent performance and training reliability in ACD.

\subsection{Yawning Titan Cyber Gym} 

% YT is highly configurable, allowing for customisable game dynamics, actions, and reward functions. 
%This tightly controlled setup was needed to ensure sufficient signal-to-noise ratio for meaningful learning. 

Yawning Titan (YT)~\citep{yawningtitan} is a well-established cyber gym providing an abstract, graph based network simulation environment for training defensive (blue) agents to defend a network by minimising the number of compromised nodes. To establish foundational insights, and to minimise variance and implementation errors in the first instance, we configured YT to simulate a linear network structure with a fixed entry node for the attacking (red) agent which follows a fixed lateral-movement strategy aiming to compromise as many nodes as possible. %Experimentation in YT is motivated by the high variability and implementation errors introduced by more complex network structures or red agent policies, making it exponentially more difficult to determine and understand the impact of different reward functions on agent performance.

% BELOW REMOVED FROM ICLR VERSION:
% As real computer networks are not only linear, often featuring many subnets and heterogeneous hosts, primarily investigating linear network structures is nevertheless a limitation of this work. In Appendix E we include experiments, continuing the key trends of our results, with a (cyclic) 20 node network allowing the red agent to take multiple routes when moving laterally from the entry node. 

%We configure YT to simulate a linear network structure with a single fixed entry node for the attacking (red) agent. The red agent operates using a fixed strategy and can only spread from an already compromised node to move laterally through the network. 
%The goal of the blue agent is to defend the network by minimising the number of compromised nodes. Conversely, the red agent aims to compromise as many nodes as possible. 

%We do not use node isolation, and -- they wont know what this is

The YT observation space comprises a vector embedding the network adjacency matrix and both the vulnerability and compromise status of each node. We set the vulnerability of each node to $1$, conservatively modelling the most powerful red agent whose attacks never fail. We create two action spaces: (1) basic – with two actions: "scan network" and "restore node", and (2) extended – which also adds "place decoy". The place decoy action is a proactive defence replicating the use of a deceptive "canary", a technique sometimes used to detect and delay attackers in real world networks. The red action space has two actions: "do nothing" and "basic attack", where the fixed red policy is to perform a basic attack 90\% of the time and do nothing otherwise (10\%).

%The reward functions are shown in Table \ref{tab:RewFuncDefinitions}.

%The key components of the MDPs defined by our YT configuration are (1) the observation space comprising a vector which embeds the network adjacency matrix, the vulnerability of each node, and whether any nodes are isolated or compromised .

%There are two blue action spaces used in our experiments, the first being the "simple action space" with just two actions: "scan network" and "restore node". The other is the "extended action space" which adds the custom action "place decoy". 

%This is a proactive action the blue agent can perform on a node which ensures that the next red basic attack action does not succeed. This action replicates the use of a deceptive "canary", a technique sometimes used to detect and delay attackers in real world networks.

%include the observation space, the action spaces for each agent, and the various reward functions being explored. 

%The observation space comprises of a vector including the network adjacency matrix, any isolated nodes, the compromised nodes and the vulnerabilities of each node. 

%In this configuration, the ability to isolate a node is not used and the vulnerabilities are all set to 1, indicating every node is vulnerable and will be compromised given a red agent attacks it. 

% New section on MiniCAGE environment
\subsection{Cyber Autonomy Gym for Experimentation}

%The Cyber Autonomy Gym for Experimentation (CAGE) challenges, and the Cyber Operations Research Gym (CybORG) they are built atop of, are some of the most the widely cited cyber gym environments utilised for ACD research~\citep{cage_cyborg_2023, vyas2023automated}. CybORG was designed specifically to enable training defensive RL agents in simulated network attack scenarios, with four online challenges held consecutively from 2021 to 2024. Each CAGE challenge increased in complexity, moving from a single agent task defending an enterprise network in challenges 1 and 2~\citep{kiely2023autonomous} to multi-agent tasks defending ad-hoc and multi-zone networks~\citep{kiely2025exploring} in challenges 3 and 4, respectively. 

Cyber Autonomy Gym for Experimentation (CAGE) 2~\citep{kiely2023autonomous} is one of the most popular single-agent ACD environments designed to enable training defensive RL agents in simulated network attack scenarios~\citep{cage_cyborg_2023, vyas2023automated}. Adding considerable complexity in contrast to YT, CAGE~2 defines an enterprise network with 3 subnets and 13 hosts in total: the user subnet with 5 hosts, the enterprise subnet with 3 hosts and an isolated defender host, and the operational subnet with 4 hosts. The network is separated by firewalls such that red agents must compromise multiple hosts to move from user subnet hosts, via the enterprise subnet, to the operational target. The observation space is a vector of 52~bits, comprising 4 bits detailing state and adversary information for each host. The action space includes 6 high-level actions (sleep, monitor, analyse, remove, restore, decoy) which are expanded to detail type and target for a total of 145 different actions.

In our experiments we use the refined CAGE~2 implementation, MiniCAGE~\citep{emerson2024cyborg}, which eliminates bugs and increases training speeds but otherwise has exactly the same environment dynamics, red agent behaviour, observation and action spaces, and network topology. Of the two red agents included in CAGE~2, we use the ``b-line'' attacker, which uses partial prior knowledge of the network to exploit the shortest path from entry node to impacting the operational target.

\subsection{Ground Truth} \label{method_ground_truth}
% \begin{itemize}

%To measure the real impact and progress of ACD agents, we need proper evaluation techniques that are robust to different reward functions.
% To understand the agent performance in a meaningful way, the evaluation method must be careful and considerate. 
To the best of our knowledge, previous work on ACD is limited to evaluating performance using the mean episodic reward and its variance over a number of rollouts. This assumes the MDP model captures the ``ground truth", and that the episodic reward is properly aligned with preferred ACD goals. However, cyber gyms are highly complicated environments which simulate both red and blue agent actions. According to the MDP framework actions are taken during discrete time steps, requiring a determined order in which red and blue actions occur. Current cyber gyms overlook this crucial detail and choose either a fixed order or prioritise actions according to some arbitrary function. 

% below is the updated figure for b
\begin{figure}[t] % Figure placement: Here, Top, or optional
    \centering % Centers the figure
    % \vspace{-3ex}
    \includegraphics[width=0.4\textwidth]{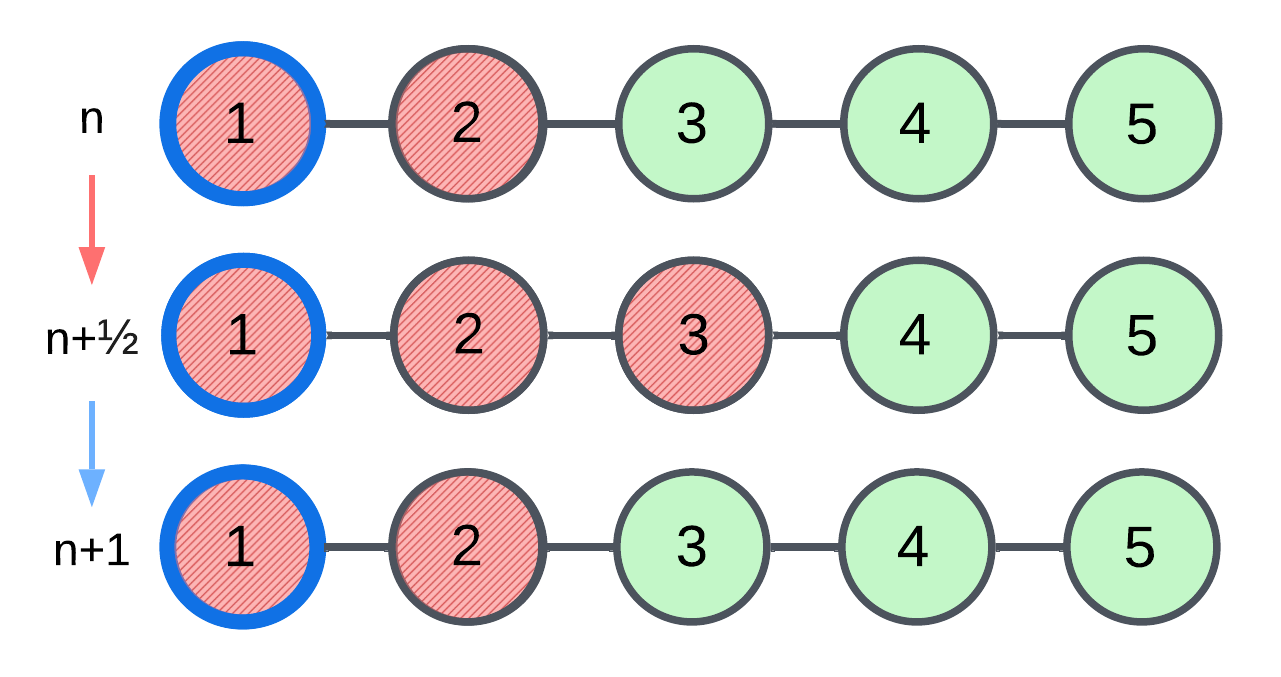} % Replace 'example-image' with your image filename
    \vspace{-1.7ex}
    \caption{One step of the YT network environment illustrating an intra-step node compromise that is concealed by standard cyber gym evaluations.}
    \label{fig:Network_diagram} % Unique label for referencing the figure
    \vspace{-1ex}
\end{figure}

Illustrated in Figure~\ref{fig:Network_diagram}, one issue with the MDP framework's requirement for discrete time steps is that the observation provided at the end of each step can omit critical network events occurring intra-step which are resolved before the reward is determined. Concretely--red agents may compromise nodes during the step, just before the blue agent removes the compromise, and this will not be reflected in the reward or observation returned to the agent. This makes it impossible for agents to reliably distinguish between states in which nodes have been compromised and those in which no compromise occurred. Consequently, prior ACD evaluation metrics fail to distinguish between agents with potentially very different ground truth behaviour.

\paragraph{Ground Truth Score ($\mathrm{Score}_\mathrm{GT}$)}
To overcome the limitations of discrete step-wise evaluation in cyber gyms we introduce the ground truth score, $\mathrm{Score}_\mathrm{GT}$, calculated as the maximum ($\mathrm{max}$) number of compromised nodes over both the intra- and end-step. In general, where $m_t^{\text{intra}}$ and $m_t^{\text{end}}$ are the intra- and end-step number of compromised nodes, respectively:
    \begin{equation}
\mathrm{Score}_{\mathrm{GT}}(t) 
= \mathrm{max} \bigl(m_t^{\text{intra}}, m_t^{\text{end}}\bigr)
\label{eq:ground_truth_score}
\end{equation}
    For Figure~\ref{fig:Network_diagram}, $\mathrm{Score}_\mathrm{GT}(\cdot) = \mathrm{max}(3, 2) = 3$ i.e., capturing the ground truth that 3 nodes were compromised during the time step. The ground truth score provides a more accurate measure of agent performance that is independent of agent order and does not depend on the reward function used during training––enabling the impact of reward on agent performance to be evaluated robustly.

%The rewards used to accumulate the episodic reward are extracted within the game abstraction: at the end of every time step. In an environment where multiple actions occur in one time step, simply looking at rewards at the end of each step means that you can obscure agent behaviour happening "mid-step". 

%We instead evaluate using a separate reward function strictly for evaluation, and that we consider the state of the network after every action rather than time step. The reward function used for evaluation is -1 per average node compromised within the time step. 

%This is subtly different from the dense reward function in table \ref{tab:RewFuncDefinitions}, as our evaluation considers the ground truth of the network state throughout the step, not just at the end. By using this reward function for evaluation, performance across different reward functions can be directly compared (see left graph in Figure \ref{fig:5 Nodes training and eval curves}). 

\vspace{-0.25ex}
\subsection{Evaluating Reliability Across Different Rewards} \label{agent reliability}
To evaluate the impact of reward function on training reliability using a single risk metric, we introduce a normalised variant of the DR measure proposed by \citet{chan2019measuring} (defined in Section~\ref{sec:background_reliability}).
To capture variability in converged performance rather than early fluctuations we restrict our application to the final 20\% of steps. For each training run $i$ we calculate the mean episodic reward $R_i$ across the final 20\% of training steps. Then, we apply mean normalisation to each run's mean episodic reward:
\vspace{-0.75ex}
    $$R'_{i} = (R_{i} - \mu)/ {\sigma}$$
Across $I$ total training runs, our normalised DR metric is calculated as the IQR over the mean normalised mean episodic rewards: 
\vspace{-0.75ex}
    $$\text{DR}' = \text{IQR}(R'_{i} \mid \forall i \in I)$$

\vspace{-1.5ex}
\subsection{Experiments}
\vspace{-0.25ex}

Our experiments evaluate the performance, risk, and reliability of the different reward functions defined in Table~\ref{tab:RewFuncDefinitions}. These reward structures are representative of both the complex, dense reward functions currently used by most cyber gyms including YT and CAGE, and an encompassing range of sparse rewards aligned with the goal of defending the network by minimising the number of compromised nodes. The sparse reward functions place fewer constraints on the optimisation objective, e.g., by avoiding numerical comparison between nodes and defensive actions, thus might enable agents to learn more effective policies. Note that we use ``positive" and ``negative" principally in reference to the goal of mitigating adversarial node compromise. The most positive outcome is that zero nodes are compromised. Correspondingly, the most negative scenario entails the complete compromise of all network nodes. See Appendix~\ref{App:PositiveAblation} for an ablation study investigating the role of reward sign versus sparsity.

%Correspondingly, we anticipate that sparse reward functions might enable agents to learn more effective policies.

In both YT and CAGE, our evaluation applies the ground truth score and reliability metrics defined in Sections~\ref{method_ground_truth} and~\ref{agent reliability}. Furthermore, upper and lower RF refer to the bounds of the average per-step ground truth score at risk across rollouts (see Section~\ref{risk}) determined by the $\alpha = 0.05$ percentile. All experiments are trained for 25 independent runs and the final policies are evaluated for 1000 episodes, resulting in a $\text{Score}_\text{GT}$, upper and lower RF, $\bar{\text{DT}}$, $\text{DR }'$ and 95\% confidence intervals (CI) for each network size, reward function, and agent order. Agent order is fixed for each corresponding training run and evaluation. We did not search for optimal hyperparameters in this work as the Stable-Baselines3 defaults (see Appendix~\ref{appendix-hyperparams}) proved sufficient in both PPO and DQN, however tuned hyperparameters may further enhance learning in any given experiment.
% Both the Stable-Baselines3 library and the Yawning Titan environment are released under the MIT License of which our work is in accordance with. 
%The hyperparameters for training are those of the default Stable-Baselines3~\citep{stable-baselines3} implementation can be seen in Appendix B. This reflects a limitation of the results; A set of tuned hyperparameters for each reward function could result in better performance across all the reward functions.
    % perfroming a sweep for the optimal hyeprparams we leave fro future work, which may ... better perf.
    % We did some informal investigation
    % A set of hyperparameter sweeps were done, resulting in hyperparameters that did not significantly differ from the default seen in the Stable-Baselines3~\citep{stable-baselines3} PPO implementation, so we retain those defaults throughout.
    % The hyperparameters for training can be seen in the appendix~\ref{appendix-hyperparams} and were tuned for each reward function separately by doing a training sweep of 150 runs between 50,000 and 200,000 steps. The sweeps were configured to use a Bayesian method and the Hyperband method which utilises early stopping to identify the best-performing configurations~\citep{hyperband}.
    Experiments were run on Intel i9 and Apple M1 and M3 Pro CPUs alongside NVIDIA RTX 4090 GPUs, requiring approximately 1478 processor days in total, including preliminary experiments and re-runs.

%(1) a "conventional"~\citep{cyborg_acd_2021, yawningtitan} dense reward function which issues a negative reward for each host that is compromised at the end of an environment step, (2) a sparse positive reward that incentivises only a network entirely free of compromised nodes, and (3) a sparse negative reward issuing a penalty any time the network is entirely compromised. 

    % By avoiding drawing any numerical equality or comparison between different nodes, which in practice would likely have different value to an attacker, the sparse rewards place fewer constraints on the optimisation objective. 

\begin{table*}[t]
  \centering
  \caption{The sparse and dense reward functions evaluated.}
  \label{tab:RewFuncDefinitions}
  % \small
  \renewcommand{\arraystretch}{1}

  \begin{tabularx}{\textwidth}{p{0.27\textwidth} p{0.27\textwidth} X}
    \toprule
    \textbf{Reward Function} & \textbf{YT Reward per time step} & \textbf{CAGE Reward per time step} \\
    \midrule
    Sparse Positive (SP) &
    +1 if no nodes compromised only. &
    +1 if no nodes compromised and red agent in user subnet. \\
    Sparse Negative (SN) &
    -1 if all nodes compromised only. &
    -1 if operational server is impacted. \\
    Sparse Positive-Negative (SPN) &
    +1 if none and -1 if all nodes compromised, respectively. &
    +1 if no nodes compromised and red agent is in user subnet, and -1 if operational server is impacted. \\
    Dense Negative (DN) &
    -1 per compromised node. &
    N/A. \\
    Complex Dense Negative (CDN) &
    Action penalties and -1 per compromised node, see Appendix~\ref{appendix_complex_rew}. &
    Standard CAGE 2 reward, see Appendix~\ref{appendix_complex_rew}. \\
    \bottomrule
  \end{tabularx}
\end{table*}

    %The sparse positive and negative reward functions are used in our experiments as they offer two different approaches to learning, the first in a way that rewards being in the goal state and the second in a way that penalises being in the worst state. Both of these are much simpler reward functions that avoid circumstances where equivalences in agent behaviour can be drawn, since they are only rewarded when either the best or worst state is reached. \\

    \vspace{-1ex}
    \subsubsection*{Yawning Titan Experiments}
    Informed by insights provided by our ground truth mechanism inSection~\ref{method_ground_truth}, we trained agents using three different orderings of red and blue actions: red then blue (standard in YT and CAGE 2), blue then red, and random. The random order performs an alternating sequence of red then blue, and blue then red, with the initial order randomised in each episode. The random order includes the worst-case for the defender where the red agent acts twice consecutively before blue can act.
    These experiments evaluate the relationship between reward structure and robustness to inter-step agent order. Prior work has shown that environment complexity and the inability to interpret agent behaviour scales rapidly as network size grows~\citep{foleyCAGEII22}. Complex network simulations obfuscate the relationship between reward function and final policy outcomes. Thus, we begin in YT with the least complex configuration: 2 nodes and 2 actions (basic); and then iteratively scale the network size up to 50 nodes before then including the proactive decoy action (extended). These experiments evaluate the impact of reward structure as both the network size and action space are scaled up. In all experiments the episode length is fixed at 100 steps and each agent is trained using PPO, one of the most widely used algorithms for training ACD agents~\citep{vyas2023automated}. To demonstrate that our findings are not specific to PPO we also perform additional experiments using DQN~\citep{mnih2015human} (see Appendix~\ref{DQN_experiments}). To ensure convergence during training, we scale the number of training steps so that for network sizes of 2, 5, 10, 20, and 50 nodes, agents are trained for 0.5, 1, 1.5, 2, and 2.5 million steps, respectively.

    %As the network size increases we also scale the number of training steps to ensure that agents always have sufficient time for their learning to stabilise. For network sizes of 2, 5, 10, 20, and 50 nodes, agents are trained for 0.5, 1, 1.5, 2, and 2.5 million steps, respectively.

    %Further experiments using DQN~\citep{mnih2015human} demonstrate our findings generalise beyond a single instance of class of DRL algorithm (see Appendix~\ref{DQN_experiments}). 

\vspace{-1ex}
\subsubsection*{CAGE Experiments}
\vspace{-0.25ex}

To explore the generalisability of our findings to non-linear network structures and expanded state-action spaces we also trained agents in the MiniCAGE environment using the set of rewards detailed in Table~\ref{tab:RewFuncDefinitions}. The episode length was fixed at 100 steps and agents were trained using both PPO and DQN for 2.5 million timesteps (see Appendix \ref{miniCAGE_dqn} for DQN results).

\begin{table*}[h]
% \vspace{-3.5ex}
\centering
\caption{PPO results in YT, for the extended action space, averaged across all network sizes and agent orders for sparse positive (SP), sparse negative (SN), sparse positive negative (SPN), dense negative (DN) and complex dense negative (CDN) reward functions.}
\label{tab:General_table_extended}
\vspace{0.1in}
\small
\setlength{\tabcolsep}{4pt}
\begin{tabular}{@{}l c c c c c c c@{}}
\toprule
%\rowcolor{yellow}
\multirow{3}{*}{ \textbf{Reward Function}} 
  & \multirow{3}{*}{\textbf{$\text{Score}_\text{GT}$}} 
  & \multicolumn{4}{c}{\textbf{Average Evaluation Reliability}}  
  & \multicolumn{2}{c}{\textbf{95\% CI}} \\ 
\cmidrule(l){3-6} \cmidrule(l){7-8}
            &              & Lower RF & Upper RF & $\bar{\text{DT}}$ (e-3) & $\text{DR}'$ 
            & LL & UL \\ 
\midrule
SP  & 2.69 & 2.46 & 2.87 & 0.11 & 0.12 & 2.02 & 3.36 \\
SN  & 10.29 & 9.00 & 10.90 & 0.09 & 0.17 & 9.10 & 11.47 \\
SPN & \textbf{2.00} & 1.82 &2.16 & 0.08 & 0.19 & 1.38 & 2.63 \\
DN  & 6.29 & 5.84 & 6.60 & 2.33 & 0.12 & 5.14 & 7.44 \\
CDN & 6.21 & 5.71 & 6.52 & 2.45 & 0.31 & 5.10 & 7.32 \\
\bottomrule
\end{tabular}
\vspace{-1ex}
\end{table*}

\vspace{-2ex}
\begin{table*}[h]
\centering
\caption{Averaged results for PPO agents trained in MiniCAGE using 4 reward functions.}
\vspace{0.1in}
\label{tab:PPO_General_table_miniCAGE}
\small
\setlength{\tabcolsep}{4pt}
\begin{tabular}{@{}l c c c c c c c@{}}
\toprule
%\rowcolor{yellow}
\multirow{3}{*}{ \textbf{Reward Function}} 
  & \multirow{3}{*}{\textbf{\begin{tabular}[c]{@{}c@{}}$\text{Score}_\text{GT}$\end{tabular}}} 
  & \multicolumn{4}{c}{\textbf{Average Evaluation Reliability}}  
  & \multicolumn{2}{c}{\textbf{95\% CI}} \\ 
\cmidrule(l){3-6} \cmidrule(l){7-8}
            &              & Lower RF & Upper RF & $\bar{\text{DT}}$ (e-3) & $\text{DR}'$
            & LL & UL \\ 
\midrule
SP   & \textbf{1.29}	&0.97	&3.11	&0.34	&0.46	&1.24	&1.34\\
SN   & 2.77	&1.85	&3.64	&0.05	&0.19	&2.66	&2.87\\
SPN  & 1.35	&0.97	&2.93	&0.36	&0.47	&1.23	&1.48\\
CDN (default CAGE rewards)    & 1.41	&1.06	&2.02	&0.55	&0.31	&1.31	&1.51\\
\bottomrule
\end{tabular}
\end{table*}

    %\begin{enumerate}
        %\item Experiments were run using a mixture of Intel i9-14900K and Mac M1 Pro processors, and NVidia 4090 GPUs.
    %    \item Total compute time of 300 G/CPU days.
    %\end{enumerate}

    % obsvsiscate the relationship between reward and agent behaviour

    % In all of our experiments, the episode length is 100 steps and each agent is trained for 1 million steps using the PPO algorithm from stablebaselines3. PPO is a state of the art DRL algorithm commonly used for training defensive autonomous agents.
    % We train 10 models per agent and average their training data for evaluation purposes. 

% \end{itemize}
    
% \end{itemize}

% \vspace{-1.25ex}
\section{Results}\label{sec:results}
\vspace{-0.5ex}
Here we present key results showing how reward structure impacts performance, risk and reliability in ACD. 
\vspace{-0.25ex}
% **** to add back in elsewhere. It's too soon here *****
%In Appendix D we provide a table outlining average agent $\text{Score}_\text{GT}$ alongside the mean episodic rewards for each individual reward function. This table allows for insight into episodic returns each reward function might achieve, and the difficulty of comparing these functions meaningfully using this metric alone, a limitation the $\text{Score}_\text{GT}$ overcomes. In the extended action space, for the red then blue order, the SN function achieves a mean episodic reward of 0, but a $\text{Score}_\text{GT}$ of 9.3. The SN function's maximum episodic reward is 0, so it has achieved it's best result yet the $\text{Score}_\text{GT}$ of 9.3 indicates that this policy acts poorly with an average of 9.3 nodes compromised per time step.

%across the two action spaces: simple and extended, network sizes ranging from 2 to 50 nodes, and 3 different agent orderings: red then blue, blue then red, and random.
% We might stick to results here, descriptive and supporting our claims but not extrapolating or analysing deeply

\subsubsection*{SP and SPN rewards perform best on average}
\vspace{-0.25ex}

Providing an overarching view of the results in YT, shown in Table 2, we consolidate the ground truth scores, risk, and training reliability of each reward function averaged across all network sizes and agent orders. The SPN reward function achieves the best scores: on average, fewer nodes are compromised than for agents trained using any other reward function. SP rewards provide the next-best performing agents, followed by DN, CDN, and finally SN rewards. All of the sparse reward functions, including SN, show significantly lower $\bar{\text{DT}}$ than the dense rewards, confirming greater training reliability across time (albeit to a low average performance for SN). 
Dense rewards exhibit $\bar{\text{DT}}$ values an  order of magnitude larger than sparse rewards, indicating significantly reduced reliability across training runs.
Across every YT configuration, as shown in Tables~4, 5, and 6, the best performing PPO agents result from either SP or SPN. This is also true for DQN agents as shown in Appendix~\ref{DQN_experiments}. Similarly in the MiniCAGE environment, as per Table~\ref{tab:PPO_General_table_miniCAGE} and Appendix~\ref{miniCAGE_dqn}, the SP reward function achieves the best $\text{Score}_\text{GT}$. Both SP and SPN rewards outperform the standard CAGE reward function in terms of $\text{Score}_\text{GT}$, and the upper 95\% confidence interval for SP is lower than the average $\text{Score}_\text{GT}$ of the standard CAGE reward function.

\subsubsection*{Performance and risk scaling with network size}
Evaluating performance as the network size increases shows how each reward function scales to larger, and therefore more realistic, state-action spaces. In YT we evaluate trained agents in networks with 2, 5, 10, 20, and 50 nodes, averaging scores over all runs and agent orders for each network size. Table~\ref{tab:agent_order_extended} shows the average ground truth performance, and worst 5\% of performances i.e., risk, for all agent orders. As network size increases, the performance and risk differences between reward functions widens. In the smallest 2 and 5 node networks, both SPN and SP yield the best agents with closely matched average performance and worst-case risks--especially in the basic action space (see Appendix~\ref{apendix_basic_action_yt_results}). In the largest two network sizes the advantages of SPN rewards are magnified, providing significantly better policies with correspondingly reduced risks. For 10 node networks there is an exception to the overall trend where SP rewards outperform SPN in the extended action space. As discussed further in Section~\ref{sec:discussion}, this is likely because in the extended action space, both SP and SPN rewards enable learning optimal strategies for defending networks when the agent order is blue then red. The results show that SP and SPN rewards not only perform best overall but also scale favourably as state-action spaces increase. Furthermore, our ablations in Appendices~\ref{App:PositiveAblation} and ~\ref{App:constant_shift_ablation} confirm this arises from sparse reward structure rather than numerical sign.
\vspace{1.25ex}

\newcommand{\tablesetupfull}{%
  \small
  \setlength{\tabcolsep}{4pt}
  \renewcommand{\arraystretch}{0.9}
}

% \vspace{-0.75ex}
\begin{table*}[h]
% \vspace{-1.75ex}
\renewcommand{\arraystretch}{0.85} % Reduce overall row height
\caption{YT PPO agent performance and risk evaluation scores for each network size in the extended action space. Results are averaged over all agent orders.}
\label{tab:across_network_results_extended}
\vspace{0.1in}
\centering
\tablesetupfull
\small
\setlength{\tabcolsep}{4pt}
\begin{tabular}{@{}p{1.3cm}cccccccccc@{}}
\toprule
\multicolumn{11}{c}{\textbf{Network sizes}} \\  
\cmidrule(l){2-11} 
\multirow{2}{*}{\makecell[c]{\textbf{Reward}\\\textbf{Function}}} 
& \multicolumn{2}{c}{\textbf{2}} & \multicolumn{2}{c}{\textbf{5}} & \multicolumn{2}{c}{\textbf{10}} & \multicolumn{2}{c}{\textbf{20}} & \multicolumn{2}{c}{\textbf{50}} \\ 
\cmidrule(lr){2-3} \cmidrule(lr){4-5} \cmidrule(lr){6-7} \cmidrule(lr){8-9} \cmidrule(lr){10-11}
& \begin{tabular}[c]{@{}c@{}}Score \\ GT\end{tabular}
& \begin{tabular}[c]{@{}c@{}}Upper \\ RF\end{tabular}
& \begin{tabular}[c]{@{}c@{}}Score \\ GT\end{tabular}
& \begin{tabular}[c]{@{}c@{}}Upper \\ RF\end{tabular}
& \begin{tabular}[c]{@{}c@{}}Score \\ GT\end{tabular}
& \begin{tabular}[c]{@{}c@{}}Upper \\ RF\end{tabular}
& \begin{tabular}[c]{@{}c@{}}Score \\ GT\end{tabular} 
& \begin{tabular}[c]{@{}c@{}}Upper \\ RF\end{tabular} 
& \begin{tabular}[c]{@{}c@{}}Score \\ GT\end{tabular} 
& \begin{tabular}[c]{@{}c@{}}Upper \\ RF\end{tabular} \\ 
\midrule
SP  & \textbf{0.60} & 0.64 & \textbf{0.62} & 0.66 & \textbf{0.63} & 0.67 & 1.87 & 1.96 & 9.75 & 10.44 \\
SN  & 1.19 & 1.26 & 3.59 & 3.82 & 7.47 & 7.92 & 12.43 & 13.28 & 26.76 & 28.20 \\
SPN & 0.92 & 0.98 & 0.97 & 1.01 & 0.85 & 0.89 & \textbf{0.69} & 0.73 & \textbf{6.58} & 7.17 \\
DN  & 0.98 & 1.03 & 1.28 & 1.41 & 3.21 & 3.42 & 8.19 & 8.45 & 17.78 & 18.70 \\
CDN & 0.85 & 0.90 & 8.73 & 9.23 & 4.03 & 4.18 & 8.46 & 8.73 & 16.06 & 17.02 \\ 
\bottomrule
\end{tabular}
\end{table*}

\vspace{-1.75ex}

\begin{table*}[h!]
\caption{YT PPO agent results for each agent action order in the extended action space, averaged over all network sizes.}
\label{tab:agent_order_extended}
\centering
\tablesetupfull
\vspace{0.1in}
\small
\setlength{\tabcolsep}{4pt}
\begin{tabular}{@{}lccccccccc@{}}
\toprule
%\rowcolor{yellow}
\multicolumn{1}{c}{\multirow{2}{*}{\makecell{\textbf{Reward}\\\textbf{Function}}}} 
& \multicolumn{3}{c}{\textbf{Red then Blue}} 
& \multicolumn{3}{c}{\textbf{Blue then Red}} 
& \multicolumn{3}{c}{\textbf{Random}} \\ 
\cmidrule(lr){2-4} \cmidrule(lr){5-7} \cmidrule(lr){8-10} 
& \multicolumn{1}{c}{$\text{Score}_\text{GT}$} 
& \multicolumn{1}{c}{Upper RF} 
& \multicolumn{1}{c}{CI UL} 
& \multicolumn{1}{c}{$\text{Score}_\text{GT}$} 
& \multicolumn{1}{c}{Upper RF} 
& \multicolumn{1}{c}{CI UL} 
& \multicolumn{1}{c}{$\text{Score}_\text{GT}$} 
& \multicolumn{1}{c}{Upper RF} 
& \multicolumn{1}{c}{CI UL} \\ 
\midrule
SP  & \textbf{0.90} & 0.96 & 0.90 & \textbf{0.27} & 0.28 & 0.72 & 6.91 & 7.38 & 8.47 \\
SN  & 9.31 & 9.95 & 10.92 & 9.01 & 9.51 & 10.72 & 12.54 & 13.23 & 12.77 \\
SPN & \textbf{0.90} & 0.96 & 0.90 & 0.61 & 0.63 & 1.23 & \textbf{4.50} & 4.89 & 5.75 \\
DN  & 4.13 & 4.31 & 5.74 & 2.98 & 3.08 & 4.17 & 11.75 & 12.42 & 12.40 \\
CDN & 5.65 & 5.80 & 5.29 & 3.99 & 4.11 & 4.14 & 13.24 & 14.13 & 12.52 \\ 
\bottomrule
\end{tabular}
\end{table*}

% \vspace{-2ex}
\subsubsection*{The impact of agent order}
Evaluating the impact of different agent orders reveals how the real-world constraints of uncertain attacker timing and dynamics could impact the performance and worst-case risks of ACD agents. Table~\ref{tab:across_network_results_extended} shows the average agent performance across all runs and network sizes in each of the three agent orders: red then blue, blue then red, and random. Continuing the trend, SP and SPN induce policies with  considerably higher performance and lower risk than the dense rewards. When agent order is randomised, the scores for all reward functions are greatly reduced, highlighting the sensitivity of DRL-based ACD agents to adversary timing. Notably, SPN significantly outperforms the other reward functions when the agent order is random--the most challenging and realistic setting in which the agent order cannot be assumed before an episode begins. Furthermore, when the agent order is blue then red and agents use the extended action space (i.e., blue can place decoys and moves before red), the average performance for SP agents reaches 0 meaning an ideally secure network with no compromised nodes during any episode. Collectively these results showcase the strong inter-relationships between reward function, action space, and performance risks when agent timing cannot be anticipated. See Appendix~\ref{episodic_rews} for the average agent $\text{Score}_\text{GT}$ alongside mean episodic rewards in YT.

\begin{table*}[h!]
\vspace{-1ex}
\caption{Agent order results for YT PPO agents trained in the 50 node network, extended action space.}
\label{tab:50_node_breakdown_results_extended}
\centering
\vspace{0.1in}
\begin{threeparttable}
\small
\setlength{\tabcolsep}{4pt}
\begin{tabular}{@{}p{1.4cm}ccccccccc@{}}
\toprule
\multirow{2}{*}[-0.9em]{\makecell{\textbf{Reward}\\\textbf{Function}}}
& \multicolumn{3}{c}{\textbf{Red then Blue}} & \multicolumn{3}{c}{\textbf{Blue then Red}} & \multicolumn{3}{c}{\textbf{Random}} \\
\cmidrule(lr){2-4} \cmidrule(lr){5-7} \cmidrule(lr){8-10} 
& \text{\small \begin{tabular}[c]{@{}c@{}}$\text{Score}_\text{GT}$\end{tabular}} 
& \text{\small \begin{tabular}[c]{@{}c@{}}Best \\ $\text{Score}_\text{GT}$\end{tabular}} 
& \text{\small \begin{tabular}[c]{@{}c@{}}No. of \\ Optimal \\ Runs (/25)\end{tabular}} 
& \text{\small \begin{tabular}[c]{@{}c@{}}$\text{Score}_\text{GT}$\end{tabular}} 
& \text{\small \begin{tabular}[c]{@{}c@{}}Best \\ $\text{Score}_\text{GT}$\end{tabular}} 
& \text{\small \begin{tabular}[c]{@{}c@{}}No. of \\ Optimal \\ Runs (/25)\end{tabular}} 
& \text{\small \begin{tabular}[c]{@{}c@{}}$\text{Score}_\text{GT}$\end{tabular}} 
& \text{\small \begin{tabular}[c]{@{}c@{}}Best \\ $\text{Score}_\text{GT}$\end{tabular}} 
& \text{\small \begin{tabular}[c]{@{}c@{}}No. of \\ Optimal \\ Runs (/25)\end{tabular}} \\
\midrule
SP  & \textbf{0.90} & 0.90 & 25 & 1.36 & 0.00 & 23 & 26.98 & 0.94 & \tnote{*} \\
SN  & 22.84 & 3.93 & 0 & 24.33 & 1.99 & 0 & 33.10 & 27.00 & \tnote{*} \\
SPN & \textbf{0.90} & 0.90 & 25 & \textbf{0.81} & 0.00 & 24 & \textbf{18.04} & 0.94 & \tnote{*} \\
DN  & 12.53 & 1.89 & 0 & 8.42 & 0.90 & 0 & 32.39 & 27.67 & \tnote{*} \\
CDN & 10.05 & 1.89 & 0 & 6.71 & 1.89 & 0 & 31.44 & 27.34 & \tnote{*}\\ 
\bottomrule
\end{tabular}
\begin{tablenotes}
\footnotesize
\item[*]{The optimal policy score is non-trivial so we do not count the number of optimal runs}
\end{tablenotes}
\vspace{-0.5ex}
\end{threeparttable}
\end{table*}
% \vspace{-2ex}

% \vspace{-1.5ex}
\section{Discussion}\label{sec:discussion}
% \vspace{-1.25ex}
%Here we discuss fully the results from Section~\ref{sec:results} and detail the conclusions we can draw about improving performance and reducing risk in DRL for ACD.
% initially inclined to separate the discussion. It should be about what we can conclude, why the results are important and what they tell us.

Empirical results for network sizes ranging from 2 to 50 nodes, irrespective of attacker timing or whether proactive decoy actions are available, confirm that SP and SPN rewards provide the best performing blue agents with minimised worst-case risks. Where the optimal scores (i.e., 0.9 for red then blue and 0 for blue then red) were computed analytically, a fine-grained evaluation of these metrics for the largest network we evaluate (shown in Table~\ref{tab:50_node_breakdown_results_extended}) further corroborates this result. Training curves for the 50 node network are in Appendix~\ref{Training_curves}. Here, both SPN and SP reward functions uniquely enable agents to learn optimal strategies which limit the attacker to very few, and even 0 in favourable conditions, compromised nodes.

In MiniCAGE, SP and SPN agents also performed best, achieving the lowest $\text{Scores}_\text{GT}$. This shows our results generalise to more realistic networks with multiple subnets and complex non-linear behaviours including hosts with different vulnerabilities. To understand why sparsely rewarded policies perform better we analyse the behaviour in Appendix~\ref{App:CAGE_policy_analysis}. While SP agents result in slightly elevated operational server impacts in comparison to the standard CAGE rewards, 0.28 vs. 0.03 average per episode; there are significantly fewer successful operational and enterprise privilege escalations, 0.33 and 1.31 vs. 8.12 and 21.80 average per episode, respectively. In addition to much lower overall privileged host access counts (22.28 vs. 34.12 average per episode), SP agents confine 21 of these (92.31\%) to user subnet hosts. This policy is much better aligned to network security objectives as user hosts have the fewest network privileges, and the smallest overall impact on operations. 

Appendix~\ref{App:CAGE_policy_analysis} also confirms that sparsely rewarded agents use the costly restore action more sparingly, and with greater focus on the user hosts, than agents trained using the standard CAGE rewards. Given the lack of numerical penalty for these actions in SP and SPN, this result highlights that sparse rewards can avoid riskier, less aligned policies which may otherwise result from incorrectly translating human domain insights into numerical rewards.
Our results demonstrate that dense rewards can suboptimally constrain the performance of ACD agents, introducing avoidable risks and reducing training reliability across runs. Furthermore, our ground truth scoring mechanism and its application in this work illustrates the importance of more granular evaluation in cyber gym environments. Many cyber gyms fail to capture important agent behaviours (e.g., including compromised intra-step nodes), obscuring crucial performance and risk differences between policies.

Dense reward functions, which are standard practice in cyber gyms, risk constraining or misaligning the performance of ACD agents and weakening the resulting security of networks they defend. More broadly, our results show that ACD agents require a reward function to provide sufficient reward signal (i.e., "can be encountered sufficiently often during training") and goal-alignment. Since dense rewards often introduce bias, sparse rewards are indicated for goal-alignment. However, sparse rewards can present exploration problems as the learning dynamics of training are highly task and algorithm dependent. The sparse positive rewards constructed here are both sparse in action-state space, and can be encountered frequently provided the uncompromised network state(s) can be identified. 

It remains for future work to understand the scaling limitations of this approach in real world networks, but the SN reward illustrates an additional challenge faced by ACD reward structures conditioned on negative outcomes for the defender. As the defensive policy improves throughout training the network becomes less frequently compromised and provides less reward signal from which to learn further improvements. 
% These findings likely have applications in many other cyber defence tasks beyond network defence, for example in web application vulnerability discovery~\citep{Lee_22, Al_Wahaibi_23}. 
These findings likely apply to other cyber security tasks beyond network defence. For example, in web application vulnerability discovery, an improving attacker increases target downtime and thus reduces reward signal per unit of wall-clock time~\citep{wendigo}.
Whilst this work is intended to advance cyber defence, a potential negative impact is its dual use by adversaries who could seek to use it for malicious purposes. As cyber gyms increase in realism, moving ACD agents closer to operational environments, it is essential to establish and empirically validate the design of effective, efficient, and risk-reducing reward functions.

\subsection{Practical Guidance for ACD Reward Design}
The results presented in this paper motivate four practical recommendations for reward function design and evaluation in ACD.
\begin{enumerate}
    \item \textbf{Adopt sparse, goal-aligned reward functions by default.}
    %Where it is possible to regularly encounter the uncompromised network state
    In this work we have shown that sparse rewards (SP and SPN) achieve the lowest $\text{Score}_\text{GT}$ in nearly every network size and agent-order in both YT and MiniCAGE. In addition, these policies have lower fixed-policy risks, show improved training reliability, and in MiniCAGE are better aligned to network security objectives. This recommendation comes with two caveats: (1) avoid sparse rewards conditioned on the negative outcome (e.g., SN) as the reward signal will vanish as the policy improves, and (2) the numerical sign does impact learning dynamics (see Appendix \ref{App:constant_shift_ablation}). SP seems to provide especially efficient learning dynamics, and greater robustness to choice of algorithm, than its constant-shifted equivalents.

%Goal-alignment alone is insufficient: SN is goal-aligned but its signal vanishes as the policy improves, while SP and SPN remain informative throughout training because the uncompromised state is encountered often. 
%Our constant-shift analysis (Appendix \ref{App:constant_shift_ablation}) further shows that even goal-aligned, frequent rewards can learn less effectively when the implied value function target is hard to learn. Under standard PPO, only the c = 0 case (where $V(s_\text{safe}) = 0$) learns reliably, as shared value/policy clipping budgets are dominated by value loss. Reward structure should therefore be evaluated jointly with the algorithm's value-learning dynamics, not in isolation.

%The default CAGE challenge 2 reward function illustrates this: it draws numerical equivalences between hosts of different causal importance, and values the restore action equivalently to enterprise/operational compromise. The resulting policies (Table 19 and 20, Appendix \ref{App:CAGE_policy_analysis}) restore enterprise hosts at the expense of user hosts, despite user compromise being causally upstream of operational impact. Sparse, goal-aligned rewards avoid this class of bias by not introducing any numerical equivalences between actions and host types.

%\item \textbf{Privileged training-time information should be exploited for evaluation and reward design.} 

\item \textbf{Look beyond episodic rewards and make use of ground truth scoring functions during evaluation.}

Our findings have revealed the presence of intra-step security-relevant agent behaviour that existing cyber gyms, and their reward functions, have not previously considered. In both YT and MiniCAGE, $\text{Score}_\text{GT}$ reveals performance and risk differences between policies that are obscured under mean episodic reward evaluation. In several configurations the relative ranking of agents is likely to change under ground truth scoring, and therefore previously reported results should be re-evaluated where possible.

%Simulator state that is unavailable at inference, such as intra-step compromise counts and true adversary position, can still shape better metrics ($\text{Score}_\text{GT}$) and inform reward functions whose induced policies remain deployable. This reframes the partial-observability gap between training and deployment as something useful for improving evaluation and reward design.

\item \textbf{Evaluate ACD agents using randomised intra-step agent ordering by default.} 

Fixed turn-based agent ordering encodes an optimistic assumption about the adversary that is unlikely to hold in operational settings. We have shown that the intra-step ordering of red and blue actions affects both $\text{Score}_\text{GT}$ and the relative ordering of reward functions in YT, and that the random ordering substantially degrades all evaluated reward functions (e.g., Table~\ref{tab:agent_order_extended}).

\item \textbf{Report risk-after-training and training reproducibility across runs.}
Our results have shown that reward function choice has a significant effect on both the worst-case performance of trained policies and the reproducibility of training, two of the most significant real-world costs in operationalising ACD agents, neither of which is captured by mean episodic rewards. We recommend reporting CVaR-based fixed-policy risk (RF), dispersion variability across runs (DR), and dispersion variability across time (DT) as introduced in Section~\ref{sec:background_reliability}. As we have demonstrated in this work, these metrics provide a useful and discriminative reward function evaluation tool in ACD settings.

%Together, these provide robust and quantitative 

    %\item \textbf{Dense rewards risk surprising policy biases in partially observable cyber environments.} 

    %\item \textbf{Reliability differences of this magnitude demand more rigorous significance evaluation.} 
    
    %The two-orders-of-magnitude $\bar{\text{DT}}$ gap between dense and sparse rewards seen in Table \ref{tab:General_table_extended} means dense-reward studies require substantially more independent runs to characterise the performance distribution. The 25 seeds per configuration used in this work are necessary to characterise this variability; published ACD results based on fewer runs may not be reliably reproducible.
\end{enumerate}

\vspace{-1.5ex}
\section{Related Work}
% \vspace{-1.5ex}

% Let's try from least to most

% The signficance of reward functions, and their formulation, at large and in general
% seque to reward shaping
Reward functions, and how best to formulate them, have been widely discussed in relation to the emergence of intelligent behaviour within the RL framework~\citep{silver21rewardsenough, vamplew22rewnotenough}. Many real-world RL applications including robotics~\citep{dorigo99} and video games~\citep{openai2019dota2largescale,song2019doomrl} utilise reward shaping to address sample inefficiency, aiming to guide learning towards useful policies by incorporating domain knowledge to reduce the learning problem difficulty. Reward shaping also arises when gradient-based methods are used to augment extrinsic rewards, such as the adaptive utilisation of a reward shaping function~\citep{hu20shapereward}, or to provide ``intrinsic motivation'' towards uncertainty-reducing actions~\citep{pathak2017curiosity}. Nevertheless, a requirement for policy invariance is that reward shaping functions must apply the difference of an arbitrary potential function between successive states~\citep{ng99PBRS}. Any other reward transformation may bias resulting trained policies away from the optimal solution~\citep{riedmiller18a}. This work establishes for the first time, with implications for widely-used cyber gyms, the performance, risk, and training efficiency implications of reward function design in ACD. Prior work has sought to benchmark RL algorithm performance~\citep{pmlr-v48-duan16}, assess algorithm reliability~\citep{henderson2018deep}, and measure policy reliability during and after training~\citep{chan2019measuring}. Whilst these methods help to evaluate an agent trained using a specific reward function, comparing multiple reward functions remains challenging, and often task-specific, because episodic-reward-based evaluation crucially lacks an external frame of reference. Safe RL methods use CVaR as a training-time constraint to optimise risk-sensitive policies~\citep{10444044, ying2022towards, kim2022efficient}. In this work we apply it to characterise the worst-case behaviour induced by different reward structures, but using CVaR as a training-time constraint in ACD remains a promising direction for future work. DRL has been used for many real-world cyber security tasks including alert prioritisation~\citep{Tong_2020}, language model "jailbreak" prompt optimisation~\citep{chen2024when}, fuzzing compilers~\citep{Li_Liu_2022}, finding web application vulnerabilities~\citep{Lee_22, Al_Wahaibi_23}, finding cache timing attacks~\citep{Luo_Mulong_23}, and overcoming hardware trojan detection methods~\citep{Gohil_22}. 

In ACD, the CAGE~4 benchmark~\citep{cage4} extends CAGE to a multi-agent enterprise network, and~\citet{singh2024hierarchical} propose a hierarchical MARL architecture for it; both use CAGE's default dense rewards and evaluate on episodic reward, leaving reward structure's effect on performance and risk unexamined. \citet{symes2023entity} address topology generalisation in YT via entity-based transformer policies, orthogonal to reward design, while \citet{purves24} add a causal structure to the YT PPO reward model and compare two reward variants; their evaluation does not address worst-case risk, training reliability, agent ordering, or scaling. In contrast, we evaluate sparse and dense rewards across two gyms, two DRL algorithms, and varying network sizes and agent orderings, under a reward-independent metric with explicit fixed-policy risk and training reliability. For a broader survey on RL-based ACD we refer readers to \citet{vyas2023automated}. The closest previous work~\citep{batesCAGE2_23} investigates 4 different reward shaping approaches (normalised, linearly scaled, non-linear scaling, and curiosity-based exploration~\citep{pathak2017curiosity}) in the CAGE environment. In contrast to this work, their results are inconclusive, policies are evaluated using only episodic rewards, and no consideration is given to DRL algorithm, agent order, policy risks, training reliability, or the effects of scaling network size or action spaces.

%% -- below reads well but it's not related work, more marketing and positioning for this paper (e.g., it has no references)
%In this paper, we illustrate and address the need for improved ACD evaluation metrics that more accurately reflect the performance and risks of agents in cyber gyms. 
%Flawed evaluation approaches have been overlooked by many cyber gyms in favour of mean episodic rewards, creating avoidable risks both in terms of agent performance (i.e., weaker cyber security) and failing to measure ACD performance during worst-case scenarios. We introduce and applying the ground truth scoring mechanism, accurately capturing previously hidden policy differences and enabling the comparison of arbitrary reward functions. Our comparative analysis of reward functions sheds light on their effectiveness in increasingly complex cyber simulations and highlights the shortcomings of dense rewards.

% TODO: Check out \begin{enumerate}

%     \item Jordan et al., Evaluating the Performance of Reinforcement Learning Algorithms, ICML 2020.~\citep{jordanCCZT20i}

%     \item Lam et al., Risk-Aware Reinforcement Learning with Coherent Risk Measures, ICML 2023.~\citep{lam2023riskaware}

%     \item NeurIPS 2020 paper "Learning to Utilize Shaping Rewards: A New Approach of Reward Shaping"~\citep{hu20shapereward}

% \end{enumerate}
\vspace{-1.5ex}
\section{Conclusion}
% \vspace{-1.5ex}

In this work we introduce a novel ground truth scoring method and address a key shortcoming of cyber gyms: neglecting intra-step node compromises when evaluating agent performance. This work enables a more accurate, risk-aware, and comprehensive evaluation of ACD policies, independent of the training reward structure or agent-timings. Through extensive experiments in YT and CAGE, two well-established cyber gyms, we show that agents trained with simpler, sparse reward functions outperform those trained on conventional dense rewards and maintain higher reliability across increasing network sizes. Notably, our SPN reward function yields policies with significantly fewer compromised nodes in worst-case scenarios, especially when attacker timing is randomised (i.e., the most realistic setting). Our findings underscore the great importance of reward functions and their relationship to risk and goal alignment in cyber environments. Lastly, we have highlighted the complex inter-relationships between reward functions, action spaces, network size, and attacker timings, relating them to the ground truth performance of ACD agents. 

%In this work we evaluate whether sparse reward functions could enable more effective ACD agents to be trained. We introduce a ground truth evaluation score, and three types of sparse reward function, which when implemented in a popular cyber gym show that sparse rewards can provide both more effective ACD agents and increased training reliability. Furthermore, our results demonstrate the complex inter-relationship between reward structure, action space, and MDP definitions in the context of cyber defence.

\section{Limitations}
\label{sec:limitations}
We highlight the following limitations of our work.

\subsubsection*{Simulation-to-reality gap}
Our findings are derived from YT and CAGE, which abstract away substantial complexity of real operational networks: real adversaries are not turn-based, may exhibit adaptive and previously unseen behaviours. ACD environments capable of modelling these properties are still an open research challenge, and validating our conclusions in them as they mature is important future work. Relatedly, $\text{Score}_\text{GT}$ relies on privileged simulator state (intra- and end-step compromise counts) that is unavailable in production due to partial observability and adversarial interference. We therefore position it as an evaluation and reward-design tool intended for use during training, rather than a signal available at inference time.
 
\subsubsection*{Scope across algorithms and reward configurations.} Although we evaluate both policy-gradient (PPO) and value-based (DQN) algorithms across two environments, multiple network sizes, and several agent orderings, there are other RL algorithms and dense reward configurations that could be explored. The dense rewards we study (DN, CDN) are representative of those used in current cyber gyms and ACD competitions but do not exhaust the space of shaping schemes. Policy-invariant shaping requires a potential difference between states~\citep{ng99PBRS}, and whether such a function can be instantiated under the partial observability of ACD remains an open question. More broadly, sparse rewards in ACD work well in part because each episode begins in (or near) the goal state of an uncompromised network; this property may not hold in other safety-critical RL domains, and we do not claim that sparse rewards are universally superior to dense ones.

\clearpage
\section*{Impact Statement}
%The main focus of this work is constructing more effective autonomous cyber defence agents.
%In the introduction, we motivate this paper by discussing the need for effective autonomous cyber defence considering society’s ever-growing dependence on digital systems and the increasing automation of attacks

This work aims to improve the rigour with which autonomous cyber defence (ACD) agents are trained and evaluated, and thus to bring deployable ACD closer to a standard at which it can responsibly augment human network defenders against the rising scale and sophistication of attacks on critical digital infrastructure.

The methods and metrics proposed here are defensive in orientation and operate over abstract network simulations rather than real systems or exploits. They do not, in our judgement, provide a meaningful uplift to an adversary seeking to attack real-world networks. Nevertheless, as cyber gyms increase in fidelity and approach operational environments, the responsibility to empirically validate the performance, reliability, and risk profile of ACD agents prior to deployment grows proportionately. We view the evaluation methods in this work as a prerequisite rather than an enhancement to that broader effort.

%. Furthermore, the environments we adapt for our experimental method are abstract representations and, even should they be adapted for offensive purposes, will not yield agents capable of attacking real-world networks. We nevertheless acknowledge the dual-use nature of ACD research: methods that improve autonomous decision-making in adversarial settings may in principle inform offensive tools. Our contributions are framed around defensive goal alignment, with $\text{Scores}_\text{GT}$ designed to evaluate defender performance specifically.

% \section{Use of Generative AI}
% In this paper we made occasional use of LLMs, namely ChatGPT, to help overcome writers block and for help with improving visual aspects of our tables.

% Evaluation, Reward functions, 

%% -- No acks for blind review
\section*{Acknowledgements}
This research was funded and supported by the UK National Cyber Security Centre (NCSC) and the Defence Science and Technology Laboratory (DSTL), supporting the Autonomous Resilient Cyber Defence (ARCD) project within the DSTL Cyber Defence Enhancement programme.

% \newpage

\bibliography{bibliography}
\bibliographystyle{icml2026}

%%%%%%%%%%%%%%%% Appendix here %%%%%%%%%%%%%%%%%%%%
% icml say you can have a single or double column appendix.

\newpage
\onecolumn
\appendix

\section{Complex Dense Negative Reward Function} \label{appendix_complex_rew}
The complex dense negative (CDN) reward function is charitably (i.e., we favourably interpret the spirit of these rewards rather than focusing on specific flaws) derived from the heavily shaped reward functions used by several of the most popular cyber gyms including CAGE's CybORG, PrimATE, and Yawning Titan (YT)~\citep{cage_cyborg_2023, yawningtitan}. A typical "real world" CDN-type reward function, combining both negative penalties and positive rewards for various blue agent actions and environment states, is taken from the YT GitHub repository and partially described in Table~\ref{tab:reward_terms} below. The full YT reward function includes additional rewards with detailed nuances and caveats not shown here for clarity of presentation. Although somewhat intuitive, the specific YT reward values are arbitrary and no clear justification is provided for the magnitudes and inevitable numerical equivalences assigned. 

%(I have a meeting now. Have to leave this half finished)
%^for specific blue actions (e.g., ) and rewards for removing red nodes and reducing the vulnerability of nodes. 

%In the YT "standard rewards" function, there are penalties related to high numbers of compromised nodes in the network, but these statements include some actions not used in our work such as "isolate node", "connect node" and "make node safe". \\

We constructed the CDN and DN reward functions to charitably represent, with decreasing complexity and shaping, the reward functions found in leading cyber gyms. Some rewards were considerably simplified, for example the penalties for compromised node states, and others were omitted entirely because our experimental designs exclude the actions altogether. Our newly introduced decoy action was assigned an arbitrary penalty based on the insight that restoring a node entirely would clearly be more disruptive than temporarily disturbing one node service. 

\newcolumntype{C}[1]{>{\centering\arraybackslash}p{#1}}

% -----------------------------------------
\begin{table*}[h!]
\renewcommand{\arraystretch}{0.98}      % compact rows
\caption{Action- and state-level shaping terms for the \textit{YT}, \textit{CDN}, and \textit{DN} reward functions.}
\vspace{0.1in}
\centering
\label{tab:reward_terms}
\small
\setlength{\tabcolsep}{4pt}
\begin{tabular}{@{}p{3.0cm}*{4}{C{4cm}}@{}}
\toprule
 & \multicolumn{3}{c}{\textbf{Reward Function}} \\
\cmidrule(lr){2-4}
 & \textbf{YT} & \textbf{CDN} & \textbf{DN} \\
\midrule
\multicolumn{4}{l}{\textbf{Actions}}\\
\midrule
Reduce Vulnerability & $-0.5$ & — & — \\
Restore Node         & $-1$   & $-0.5$ & $0$ \\
Make Node Safe       & $-0.5$ & — & — \\
Scan Network         & $0$    & $0$ & $0$ \\
Isolate Node         & $-10$  & — & — \\
Connect Node         & $0$    & — & — \\
Add Deceptive Node   & $-8$   & — & — \\
Place Decoy          & —      & $-0.25$ & $0$ \\
Do Nothing           & $+0.5$ & $-0.1$ & $0$ \\
\midrule
\multicolumn{4}{l}{\textbf{States}}\\
\midrule
\makecell[tl]{Network compromise\\ / vulnerability}   % ← top-left aligned
  & \makecell[t]{%
      \textit{$>$30\% nodes compromised:}\\
      $-1$ per compromised node\\[0.3em]
      \textit{Vulnerability reduced:}\\
      $+4 \times$ reduction}
  & \makecell[t]{$-1$ per\\ compromised node}
  & \makecell[t]{$-1$ per\\ compromised node} \\
\bottomrule
\end{tabular}
\end{table*}
 
\newpage
\section{Hyperparameters for training} \label{appendix-hyperparams}
 
Here we present the hyperparameters used for training in both the YT and MiniCAGE environments.
 
\begin{table}[!h]
    \centering
    % \small
    \caption{Hyperparameters for PPO models}
    \begin{tabular}{lc}
        \toprule
        \textbf{Hyperparameter} & \textbf{Value} \\
        \midrule
        Learning Rate & $3 \times 10^{-4}$ \\
        Number of Hidden Layers & 2 \\
        Hidden Layer Size & 64 \\
        GAE Lambda & 0.95 \\
        Clip Range & 0.2 \\
        Gamma & 0.99 \\
        Value Function Coefficient & 0.5 \\
        Number of Epochs & 10 \\
        Batch Size & 64 \\
        \bottomrule
    \end{tabular}
\end{table}
 
\begin{table}[!h]
    \centering
    % \small
    \caption{Hyperparameters for DQN models, using the default Stable-Baselines3 hyperparameters with the exception of the buffer size (changed from $1e^{6}$ to $200,000$) and final epsilon (changed from $0.05$ to $0.005$)}
    \vspace{0.1in}
    \begin{tabular}{lc}
        \toprule
        \textbf{Hyperparameter} & \textbf{Value} \\
        \midrule
        Learning Rate & $1 \times 10^{-4}$ \\
        Batch Size & 32 \\
        Gamma & 0.99 \\
        Train/Update Frequency & 4 \\
        Buffer Size & 200,000 \\
        Exploration Initial Epsilon & 1 \\
        Exploration Final Epsilon & 0.005 \\
        \bottomrule
    \end{tabular}
\end{table}

%%%%%%%%%%%%%%%%%%%%%%%%%%%%%%%%%%%%%%%%%%%%%%%%%%%%%%%%%%%%
\newpage
\section{Basic Action Space Results in Yawning Titan}
\label{apendix_basic_action_yt_results}

Here we detail our YT results using the basic action space comprising (1) scan network and (2) restore node.

\begin{table*}[h!]
\centering
\caption{PPO results (Basic Action Space) averaged across all network sizes and agent orders for sparse positive (SP), sparse negative (SN), sparse positive negative (SPN), dense negative (DN) and complex dense negative (CDN) reward functions.}
\label{tab:General_table_basic}
\vspace{0.1in}
\small
\setlength{\tabcolsep}{4pt}
\begin{tabular}{@{}l c c c c c c c@{}}
\toprule
%\rowcolor{yellow}
\multirow{3}{*}{ \textbf{Reward Function}} 
  & \multirow{3}{*}{\textbf{$\text{Score}_\text{GT}$}} 
  & \multicolumn{4}{c}{\textbf{Average Evaluation Reliability}}  
  & \multicolumn{2}{c}{\textbf{95\% CI}} \\ 
\cmidrule(l){3-6} \cmidrule(l){7-8}
            &              & Lower RF & Upper RF & $\bar{\text{DT}}$ (e-3) & $\text{DR}'$ 
            & LL & UL \\ 
\midrule
SP  & 4.58 & 4.16 & 4.88 & 0.07 & 0.19 & 3.83 & 5.32 \\
SN  & 9.92 & 8.90 & 10.69 & 0.05 & 0.21 & 8.82 & 11.03 \\
SPN & 1.97 & 1.75 & 2.42 & 0.09 & 0.26 & 1.47 & 2.46 \\
DN  & 5.84 & 5.42 & 6.16 & 2.98 & 0.29 & 4.73 & 6.04 \\
CDN & 6.03 & 5.61 & 6.37 & 2.90 & 0.39 & 5.21 & 6.86 \\
\bottomrule
\end{tabular}
\vspace{-2ex}
\end{table*}

\begin{table*}[h!]
\renewcommand{\arraystretch}{0.85} % Reduce overall row height
\caption{PPO agent performance and risk evaluation scores across network sizes (Basic Action Space). Results are averaged over all agent orders for each of the 5 reward functions.}
\label{tab:across_network_results_basic}
\vspace{0.1in}
\centering
\small
\setlength{\tabcolsep}{4pt}
\begin{tabular}{@{}p{1.5cm}cccccccccc@{}}
\toprule
% \rowcolor{yellow}
\multicolumn{11}{c}{\textbf{Evaluation across network sizes}} \\  
\cmidrule(l){2-11} 
\multirow{2}{*}{\makecell{\textbf{Reward}\\\textbf{Function}}} 
& \multicolumn{2}{c}{\textbf{2}} & \multicolumn{2}{c}{\textbf{5}} & \multicolumn{2}{c}{\textbf{10}} & \multicolumn{2}{c}{\textbf{20}} & \multicolumn{2}{c}{\textbf{50}} \\ 
\cmidrule(lr){2-3} \cmidrule(lr){4-5} \cmidrule(lr){6-7} \cmidrule(lr){8-9} \cmidrule(lr){10-11}
& \begin{tabular}[c]{@{}c@{}}Score \\ GT\end{tabular}
& \begin{tabular}[c]{@{}c@{}}Upper \\ RF\end{tabular}
& \begin{tabular}[c]{@{}c@{}}Score \\ GT\end{tabular}
& \begin{tabular}[c]{@{}c@{}}Upper \\ RF\end{tabular}
& \begin{tabular}[c]{@{}c@{}}Score \\ GT\end{tabular}
& \begin{tabular}[c]{@{}c@{}}Upper \\ RF\end{tabular}
& \begin{tabular}[c]{@{}c@{}}Score \\ GT\end{tabular}
& \begin{tabular}[c]{@{}c@{}}Upper \\ RF\end{tabular}
& \begin{tabular}[c]{@{}c@{}}Score \\ GT\end{tabular}
& \begin{tabular}[c]{@{}c@{}}Upper \\ RF\end{tabular} \\ 
\midrule
SP  & 1.05 & 1.11 & 1.10 & 1.17 & 2.05 & 2.22 & 3.88 & 4.01 & 14.81 & 15.32 \\
SN  & 1.23 & 1.29 & 3.72 & 3.90 & 6.87 & 7.22 & 12.08 & 13.27 & 25.72 & 26.16 \\
SPN & 1.05 & 1.11 & 1.23 & 1.97 & 1.05 & 1.11 & 1.23 & 1.31 & 5.30 & 5.88 \\
DN  & 1.16 & 1.22 & 1.26 & 1.32 & 3.39 & 3.52 & 8.26 & 8.51 & 15.13 & 16.10 \\
CDN & 1.31 & 1.36 & 8.03 & 8.28 & 4.23 & 4.35 & 7.58 & 7.83 & 14.70 & 15.84 \\
\bottomrule
\end{tabular}
\end{table*}

\begin{table*}[h!]
\caption{PPO agent results for each agent action order combination (Basic Action Space), averaged over all network sizes, for each of the 5 reward functions. CI UL is the upper limit of the 95\% confidence interval.}
\label{tab:agent_order_basic}
\centering
\vspace{0.1in}
\small
\setlength{\tabcolsep}{4pt}
\begin{tabular}{@{}lccccccccc@{}}
\toprule
% \rowcolor{yellow}
\multicolumn{1}{c}{\multirow{2}{*}{\makecell{\textbf{Reward}\\\textbf{Function}}}} 
& \multicolumn{3}{c}{\textbf{Red then Blue}} 
& \multicolumn{3}{c}{\textbf{Blue then Red}} 
& \multicolumn{3}{c}{\textbf{Random}} \\ 
\cmidrule(lr){2-4} \cmidrule(lr){5-7} \cmidrule(lr){8-10} 
& \multicolumn{1}{c}{$\text{Score}_\text{GT}$} 
& \multicolumn{1}{c}{Upper RF} 
& \multicolumn{1}{c}{CI UL} 
& \multicolumn{1}{c}{$\text{Score}_\text{GT}$} 
& \multicolumn{1}{c}{Upper RF} 
& \multicolumn{1}{c}{CI UL} 
& \multicolumn{1}{c}{$\text{Score}_\text{GT}$} 
& \multicolumn{1}{c}{Upper RF} 
& \multicolumn{1}{c}{CI UL} \\ 
\midrule
SP  & 0.90 & 0.96 & 0.90 & 5.11 & 4.99 & 6.64 & 7.73 & 8.35 & 8.43 \\
SN  & 8.02 & 8.71 & 9.69 & 9.55 & 9.43 & 10.93 & 12.20 & 12.96 & 12.46 \\
SPN & 0.90 & 0.96 & 0.90 & 1.11 & 1.54 & 1.28 & 3.90 & 4.34 & 5.21 \\
DN  & 3.22 & 3.30 & 3.93 & 3.10 & 3.10 & 4.09 & 11.19 & 12.00 & 11.76 \\
CDN & 3.80 & 3.87 & 4.24 & 4.73 & 4.86 & 4.36 & 12.97 & 13.87 & 11.98 \\
\bottomrule
\end{tabular}
\end{table*}
 
\begin{table*}[h!]
\caption{Detailed PPO results for agents trained in the 50 node network (Basic Action Space).}
\label{tab:50_node_breakdown_results_basic}
\centering
\vspace{0.1in}
\small
\setlength{\tabcolsep}{4pt}
\begin{tabular}{@{}p{1.4cm}ccccccccc@{}}
\toprule
% \rowcolor{yellow}
& \multicolumn{9}{c}{\textbf{50 Node Network Evaluation}} \\ 
\cmidrule(l){2-10} 
 
& \multicolumn{3}{c}{\textbf{Red then Blue}} & \multicolumn{3}{c}{\textbf{Blue then Red}} & \multicolumn{3}{c}{\textbf{Random}} \\
\cmidrule(lr){2-4} \cmidrule(lr){5-7} \cmidrule(lr){8-10} 
\multirow{-2}{*}{\makecell{\textbf{Reward}\\\textbf{Function}}} 
& \text{\small \begin{tabular}[c]{@{}c@{}}$\text{Score}_\text{GT}$\end{tabular}} 
& \text{\small \begin{tabular}[c]{@{}c@{}}Best \\ $\text{Score}_\text{GT}$\end{tabular}} 
& \text{\small \begin{tabular}[c]{@{}c@{}}No. of \\ Optimal \\ Runs (/25)\end{tabular}} 
& \text{\small \begin{tabular}[c]{@{}c@{}}$\text{Score}_\text{GT}$\end{tabular}} 
& \text{\small \begin{tabular}[c]{@{}c@{}}Best \\ $\text{Score}_\text{GT}$\end{tabular}} 
& \text{\small \begin{tabular}[c]{@{}c@{}}No. of \\ Optimal \\ Runs (/25)\end{tabular}} 
& \text{\small \begin{tabular}[c]{@{}c@{}}$\text{Score}_\text{GT}$\end{tabular}} 
& \text{\small \begin{tabular}[c]{@{}c@{}}Best \\ $\text{Score}_\text{GT}$\end{tabular}} 
& \text{\small \begin{tabular}[c]{@{}c@{}}No. of \\ Optimal \\ Runs (/25)\end{tabular}} \\
\midrule
SP  & 0.90 & 0.90 & 25 & 12.68 & 0.90 & 2 & 30.86 & 25.02 & 0 \\
SN  & 19.72 & 1.89 & 0 & 26.48 & 1.89 & 0 & 30.97 & 27.06 & 0 \\
SPN & 0.90 & 0.90 & 25 & 14.31 & 0.90 & 3 & 13.57 & 1.34 & 0 \\
DN  & 7.68 & 0.90 & 2 & 8.30 & 1.89 & 0 & 29.40 & 24.41 & 0 \\
CDN & 6.36 & 0.90 & 4 & 8.77 & 1.89 & 0 & 28.96 & 27.07 & 0 \\
\bottomrule
\end{tabular}
\end{table*}

%%%%%%%%%%%%%%%%%%%%%%%%%%%%%%%%%%%%%%%%%%%%%%%%%%%%%%%%%%%%
\newpage
\section{DQN Results in Yawning Titan} \label{DQN_experiments}
Here are results for the extended action space trained using the DQN algorithm.

% These results prioritise the experimental set up where the pro-active policy of decoying a node is able to be learned. This is a more complicated task than seen in the simple action space where a reactive policy of restoring nodes is the best outcome.

% Just say what it is.

\begin{table*}[h!]
\centering
\caption{DQN results averaged over network sizes 2 to 50, and all agent orders, for all 5 reward functions: sparse positive (SP), sparse negative (SN), sparse positive negative (SPN), dense negative (DN) and complex dense negative (CDN) and use the extended action space.}
\vspace{0.1in}
\label{tab:DQN_General_table_new}
\small
\setlength{\tabcolsep}{4pt}
\begin{tabular}{@{}l c c c c c c c@{}}
\toprule
\multirow{3}{*}{\makecell[l]{\textbf{Reward}\\\textbf{Function}}} 
  & \multirow{3}{*}{\textbf{\begin{tabular}[c]{@{}c@{}}$\text{Score}_\text{GT}$\end{tabular}}} 
  & \multicolumn{4}{c}{\textbf{Average Evaluation Reliability}}  
  & \multicolumn{2}{c}{\textbf{95\% CI}} \\ 
\cmidrule(l){3-6} \cmidrule(l){7-8}
            &              & Lower RF & Upper RF & $\bar{\text{DT}}$ (e-3) & $\text{DR}'$
            & LL & UL \\ 
\midrule
% \multicolumn{6}{l}{\textbf{Extended Action Space}} \\
% \midrule
SP   & 1.48 & 0.88 & 2.38 & 0.70 & 0.44 & 1.03 & 1.93\\
SN   & 5.98 & 4.34 & 7.16 & 0.17 & 0.38 & 3.11 & 8.86\\
SPN  & 1.12 & 0.78 & 1.50 & 0.75 & 0.44 & 0.56 & 1.69\\
DN   & 3.80 & 3.16 & 4.22 & 5.83 & 0.46 & 2.81 & 4.78\\
CDN  & 5.44 & 4.83 & 5.91 & 4.17 & 0.43 & 4.27 & 6.62\\
\bottomrule
\end{tabular}
\end{table*}

\begin{table*}[h!]
\renewcommand{\arraystretch}{0.85} % Reduce overall row height
\caption{DQN performance and risk scores as network size increases. Results are averaged over all agent orders for all 5 reward functions: sparse positive (SP), sparse negative (SN), sparse positive negative (SPN), dense negative (DN) and complex dense negative (CDN) and use the extended action space.}
\vspace{0.1in}
\centering
\label{tab:across_network_results_DQN}
\small
\setlength{\tabcolsep}{4pt}
\begin{tabular}{@{}p{1.4cm}cccccccccc@{}}
\toprule
\multicolumn{11}{c}{\textbf{Evaluation across network sizes}} \\  
\cmidrule(l){2-11} 
\multirow{2}{*}{\makecell{\textbf{Reward}\\\textbf{Function}}} 
& \multicolumn{2}{c}{\textbf{2}} & \multicolumn{2}{c}{\textbf{5}} & \multicolumn{2}{c}{\textbf{10}} & \multicolumn{2}{c}{\textbf{20}} & \multicolumn{2}{c}{\textbf{50}} \\ 
\cmidrule(lr){2-3} \cmidrule(lr){4-5} \cmidrule(lr){6-7} \cmidrule(lr){8-9} \cmidrule(lr){10-11}
& \begin{tabular}[c]{@{}c@{}}Score \\ GT\end{tabular} 
& \begin{tabular}[c]{@{}c@{}}Upper \\ RF\end{tabular} 
& \begin{tabular}[c]{@{}c@{}}Score \\ GT\end{tabular}
& \begin{tabular}[c]{@{}c@{}}Upper \\ RF\end{tabular}
& \begin{tabular}[c]{@{}c@{}}Score \\ GT\end{tabular}
& \begin{tabular}[c]{@{}c@{}}Upper \\ RF\end{tabular}
& \begin{tabular}[c]{@{}c@{}}Score \\ GT\end{tabular} 
& \begin{tabular}[c]{@{}c@{}}Upper \\ RF\end{tabular} 
& \begin{tabular}[c]{@{}c@{}}Score \\ GT\end{tabular} 
& \begin{tabular}[c]{@{}c@{}}Upper \\ RF\end{tabular} \\ 
\midrule
% \multicolumn{11}{l}{\textbf{Extended Action Space}} \\ 
% \midrule
SP & 0.71	&0.78&	0.71	&0.78	&0.95	&1.13&	2.28&	3.72& 2.76&	5.51\\
SN & 0.98&	1.06&	1.99	&2.35&	2.77&	3.36&	8.81&	10.65&	15.37	&18.39 \\
SPN &0.72&	0.78&	0.71&	0.78&	0.78&	0.89&	0.73&	0.96&	2.68	&4.07\\
DN  &0.69&	0.76&	1.01&	1.11&	1.25&	1.42&	4.88&	5.43&	11.14&	12.40 \\
CDN &0.69&	0.76&	0.92&	0.99&	1.11&	1.29&	3.86&	4.28&	20.63	&22.23 \\ \bottomrule
\end{tabular}
\end{table*}

\begin{table*}[h!]
\caption{DQN results for all agent order combinations, averaged over all network sizes, for all 5 reward functions: sparse positive (SP), sparse negative (SN), sparse positive negative (SPN), dense negative (DN) and complex dense negative (CDN) and use the extended action space. CI UL is the upper limit of the 95\% confidence interval.}
\centering % Center the entire table
  \vspace{0.1in}
\label{tab:DQN results across order}
\small
\setlength{\tabcolsep}{4pt}
\begin{tabular}{@{}lccccccccc@{}}
\toprule
% \rowcolor{yellow}
\multicolumn{1}{c}{\multirow{2}{*}{\makecell{\textbf{Reward}\\\textbf{Function}}}} 
& \multicolumn{3}{c}{\textbf{Red then Blue}} 
& \multicolumn{3}{c}{\textbf{Blue then Red}} 
& \multicolumn{3}{c}{\textbf{Random}} \\ 
\cmidrule(lr){2-4} \cmidrule(lr){5-7} \cmidrule(lr){8-10} 
& \multicolumn{1}{c}{$\text{Score}_\text{GT}$} 
& \multicolumn{1}{c}{Upper RF} 
& \multicolumn{1}{c}{CI UL} 
& \multicolumn{1}{c}{$\text{Score}_\text{GT}$} 
& \multicolumn{1}{c}{Upper RF} 
& \multicolumn{1}{c}{CI UL} 
& \multicolumn{1}{c}{$\text{Score}_\text{GT}$} 
& \multicolumn{1}{c}{Upper RF} 
& \multicolumn{1}{c}{CI UL} \\ 
\midrule
% \multicolumn{7}{l}{\textbf{Extended Action Space}} \\ 
% \midrule
SP  & 0.90	& 0.96& 0.90	     &2.12& 	4.52& 	3.02 & 1.42& 	1.68 &1.88 \\
SN  & 5.82	& 6.40& 10.48    &2.68& 	3.98& 	4.54 & 9.44&   11.10 &11.55\\
SPN & 0.90	& 0.96& 0.90	     &1.25& 	2.06& 	2.86 &1.23& 	1.47 &1.31\\
DN  & 1.31	& 1.37& 1.67	 &0.77& 	0.96& 	2.03 &9.30& 	10.35&10.65\\
CDN & 1.27	& 1.33& 1.63	 &6.81& 	7.11& 	9.06 &8.24& 	9.29 &9.17\\ \bottomrule
\end{tabular}
\end{table*}
%%%%%%%%%%% Native avg rewards %%%%%%%%%%%%
\newpage

\section{Episodic Rewards with Corresponding $\text{Score}_\text{GT}$ for Yawning Titan Experiments} \label{episodic_rews}

Here we show the mean episodic rewards for each reward function, both action spaces, and all 3 agent orders. These results highlight the importance of our ground truth scoring method as it provides a common basis for comparing between agents trained using different reward functions. In addition, the poor correlation between mean episodic rewards and $\text{Score}_\text{GT}$ shows the need for better evaluation metrics in ACD. 

\begin{table*}[h!]
\caption{PPO $\text{Score}_\text{GT}$ and episodic mean rewards for all agent order combinations, averaged over all network sizes, for all 5 reward functions: sparse positive (SP), sparse negative (SN), sparse positive negative (SPN), dense negative (DN) and complex dense negative (CDN).}
\centering % Center the entire table
  \vspace{0.1in}
\label{tab:PPO results across order}
\small
\setlength{\tabcolsep}{4pt}
\begin{tabular}{@{}lcccccc@{}}
\toprule
\multicolumn{1}{c}{\multirow{2}{*}{\makecell{\textbf{Reward}\\\textbf{Function}}}} & \multicolumn{2}{c}{\textbf{Red then Blue}} & \multicolumn{2}{c}{\textbf{Blue then Red}} & \multicolumn{2}{c}{\textbf{Random}} \\ 
\cmidrule(lr){2-3} \cmidrule(lr){4-5} \cmidrule(lr){6-7} 
& \multicolumn{1}{c}{\text{\begin{tabular}[c]{@{}c@{}}$\text{Score}_\text{GT}$\end{tabular}}} & \multicolumn{1}{c}{\text{\begin{tabular}[c]{@{}c@{}}Mean Episodic \\ Reward\end{tabular}}} & \multicolumn{1}{c}{\text{\begin{tabular}[c]{@{}c@{}}$\text{Score}_\text{GT}$\end{tabular}}} & \multicolumn{1}{c}{\text{\begin{tabular}[c]{@{}c@{}}Mean Episodic \\ Reward\end{tabular}}} & \multicolumn{1}{c}{\text{\begin{tabular}[c]{@{}c@{}}$\text{Score}_\text{GT}$\end{tabular}}} & \multicolumn{1}{c}{\text{\begin{tabular}[c]{@{}c@{}}Mean Episodic \\ Reward\end{tabular}}} \\ 
\midrule
\multicolumn{7}{l}{\textbf{Basic Action Space}} \\
\midrule
SP  & 0.9 & 100.0 & 5.1 & 5.6 & 7.7 & 8.4 \\
SN  & 8.0 & -10.8 & 9.6 & -3.0 & 12.2 & -0.1 \\
SPN & 0.9 & 75.0 & 1.1 & 7.5 & 3.9 & 9.5 \\
DN  & 3.2 & -646.9 & 3.1 & -729.4 & 11.2 & -902.9 \\
CDN & 3.8 & -662.4 & 4.7 & -825.2 & 17.0 & -1001.6 \\
\midrule
\multicolumn{7}{l}{\textbf{Extended Action Space}} \\
\midrule
SP  & 0.9 & 100.0 & 0.3 & 93.5 & 6.9 & 25.1 \\
SN  & 9.3 & 0.0 & 9.0 & -2.6 & 12.5 & -0.5 \\
SPN & 0.9 & 80.0 & 0.6 & 69.0 & 4.5 & 20.6 \\
DN  & 4.1 & -492.9 & 3.0 & -532.9 & 11.8 & -1029.5 \\
CDN & 5.7 & -657.8 & 4.0 & -664.4 & 17.6 & -1100.8 \\
\bottomrule
\end{tabular}
\end{table*}

%%%%%%%%%%%%%% Appendix F Policy Analysis of YT Agents %%%%%%%%%%%%%

\clearpage
\section{Policy Analysis of YT Agents} \label{tab:YT_policy_analysis}

\begin{table}[htbp]
\centering
\caption{The average blue action counts for each set of agents in an 100 step episode, averaged across network sizes, for the extended action space. These are the result of 1000 episodes of evaluation for each of the 25 agents in each set.}
\label{tab:blue-actions-no-agent-order}
\vspace{0.1in}
\renewcommand{\arraystretch}{1.2}
\begin{tabular}{@{}lrrrrr@{}}
\toprule
\textbf{Blue actions} & \textbf{SP} & \textbf{SN} & \textbf{SPN} & \textbf{DN} & \textbf{CDN} \\
\midrule
\multicolumn{6}{@{}l}{\textbf{Action Order: Red then Blue}} \\
\midrule
Scan Network & 0 & 0.36 & 0 & 0 & 0 \\
Restore Node & 100 & 67.61 & 100 & 88.40 & 81.29 \\
Place Decoy  & 0 & 32.03 & 0 & 11.60 & 18.71 \\
\midrule
\multicolumn{6}{@{}l}{\textbf{Action Order: Blue then Red}} \\
\midrule
Scan Network & 0 & 12.24 & 0 & 0.20 & 0.98 \\
Restore Node & 0.63 & 42.96 & 32.71 & 59.84 & 41.07 \\
Place Decoy  & 99.37 & 38.07 & 67.29 & 39.96 & 57.96 \\
\midrule
\multicolumn{6}{@{}l}{\textbf{Action Order: Random}} \\
\midrule
Scan Network & 0.06 & 0.69 & 0.01 & 0.01 & 1.34 \\
Restore Node & 59.07 & 82.43 & 73.52 & 89.40 & 80.71 \\
Place Decoy  & 40.87 & 16.88 & 26.47 & 10.59 & 17.94 \\
\bottomrule
\end{tabular}
\end{table}

%%%%%%%%%%%%%% Appendix G Policy analysis of CAGE agents %%%%%%%%%%%%%
% Appendix section with behviour analysis section for the miniCAGE experiments
\newpage
\section{Policy analysis of CAGE agents} \label{App:CAGE_policy_analysis}

Here we show the detailed behaviour of agents trained in the MiniCAGE environment using sparse and dense rewards. 

\begin{table}[h]
\centering
\caption{In MiniCAGE, mean successful Impact action counts, and mean privilege red access counts, for each reward function. Evaluated over 1000, 100-step episodes.}
\vspace{0.1in}
\label{tab:impact-priv-access}
\renewcommand{\arraystretch}{1.1}
\small
\setlength{\tabcolsep}{4pt}
\begin{tabular}{@{}lcccc@{}}
 
\toprule
%\rowcolor{yellow}
\textbf{Reward Function} &
\begin{tabular}[c]{@{}c@{}}\textbf{Impact Op} \\ \textbf{Server Count}\end{tabular} &
\begin{tabular}[c]{@{}c@{}}\textbf{Op Server} \\ \textbf{Privilege access} \\ \textbf{Count}\end{tabular} &
\begin{tabular}[c]{@{}c@{}}\textbf{Enterprise host} \\ \textbf{Privilege access} \\ \textbf{Count}\end{tabular} &
\begin{tabular}[c]{@{}c@{}}\textbf{User host} \\ \textbf{Privilege access} \\ \textbf{Count}\end{tabular} \\
\midrule
SP & 0.28 & \textbf{0.33} & \textbf{1.31} & 20.64 \\
SN & 4.93 & 11.55 & 2.25 & \textbf{1.07} \\
SPN & 2.56 & 3.94 & 2.72 & 22.16 \\
CDN (Default CAGE Rewards) & \textbf{0.03} & 8.12 & 21.80 & 4.20 \\
\bottomrule
 
\end{tabular}
\end{table}

\begin{table}[h]
\centering
\caption{Mean blue action counts on each subnet and operational server for each reward function, evaluated over 1000 episodes (100 time steps). This table only includes the most relevant actions, with others like `analyse' not included for conciseness.}
\label{tab:blue-action-counts}
\vspace{0.1in}
\renewcommand{\arraystretch}{1.2}
\begin{tabular}{@{}lrrrc@{}}
\toprule
% \rowcolor{yellow}
\textbf{Action} & \textbf{SP} & \textbf{SN} & \textbf{SPN} &
\begin{tabular}[c]{@{}c@{}}\textbf{DN (default}\\ \textbf{CAGE rewards)}\end{tabular} \\
\midrule
Decoy -- User host & 1.83 & 6.27 & 1.86 & 2.78 \\
Decoy -- Ent host  & 5.00 & 5.07 & 5.00 & 1.94 \\
Decoy -- Op server & 0.68 & 4.24 & 0.29 & 1.01 \\
\textbf{Decoy Total} & \textbf{7.51} & \textbf{15.58} & \textbf{7.15} & \textbf{5.73} \\
\addlinespace
Remove -- User host & 13.62 & 10.21 & 8.27 & 1.12 \\
Remove -- Ent host  & 1.46 & 3.23 & 2.18 & 1.89 \\
Remove -- Op server & 0.31 & 0.85 & 0.18 & 0.33 \\
\textbf{Remove Total} & \textbf{15.39} & \textbf{14.29} & \textbf{10.63} & \textbf{3.34} \\
\addlinespace
Restore -- User host & 62.93 & 6.67 & 62.86 & 16.31 \\
Restore -- Ent host  & 3.01 & 4.76 & 5.18 & 57.36 \\
Restore -- Op server & 0.38 & 17.17 & 7.15 & 11.91 \\
\textbf{Restore Total} & \textbf{66.33} & \textbf{28.60} & \textbf{75.18} & \textbf{85.58} \\
\bottomrule
\end{tabular}
\end{table}

\section{Code Availability}
Our code is available at: 
\url{https://github.com/alan-turing-institute/Rewards-in-RL-for-ACD}.

% \section{Code Availability}

% \noindent\fbox{\parbox{0.97\linewidth}{%
% The code to reproduce all experiments in this paper is available at:\\
% \url{https://github.com/alan-turing-institute/Rewards-in-RL-for-ACD}
% }}
%%%%%%%%%%%%%% Appendix I PPO 50 node training curves %%%%%%%%%%%%%
\newpage
\section{PPO Agent Training Curves in Yawning Titan} \label{Training_curves}
To accompany the detailed 50 node network data in Table~\ref{tab:50_node_breakdown_results_extended}, here we provide the training curves for each reward function and agent order in the extended action space.

\begin{figure}[!h] % Figure placement: Here, Top, or optional
    \centering % Centers the figure
    \includegraphics[width=0.7\textwidth]{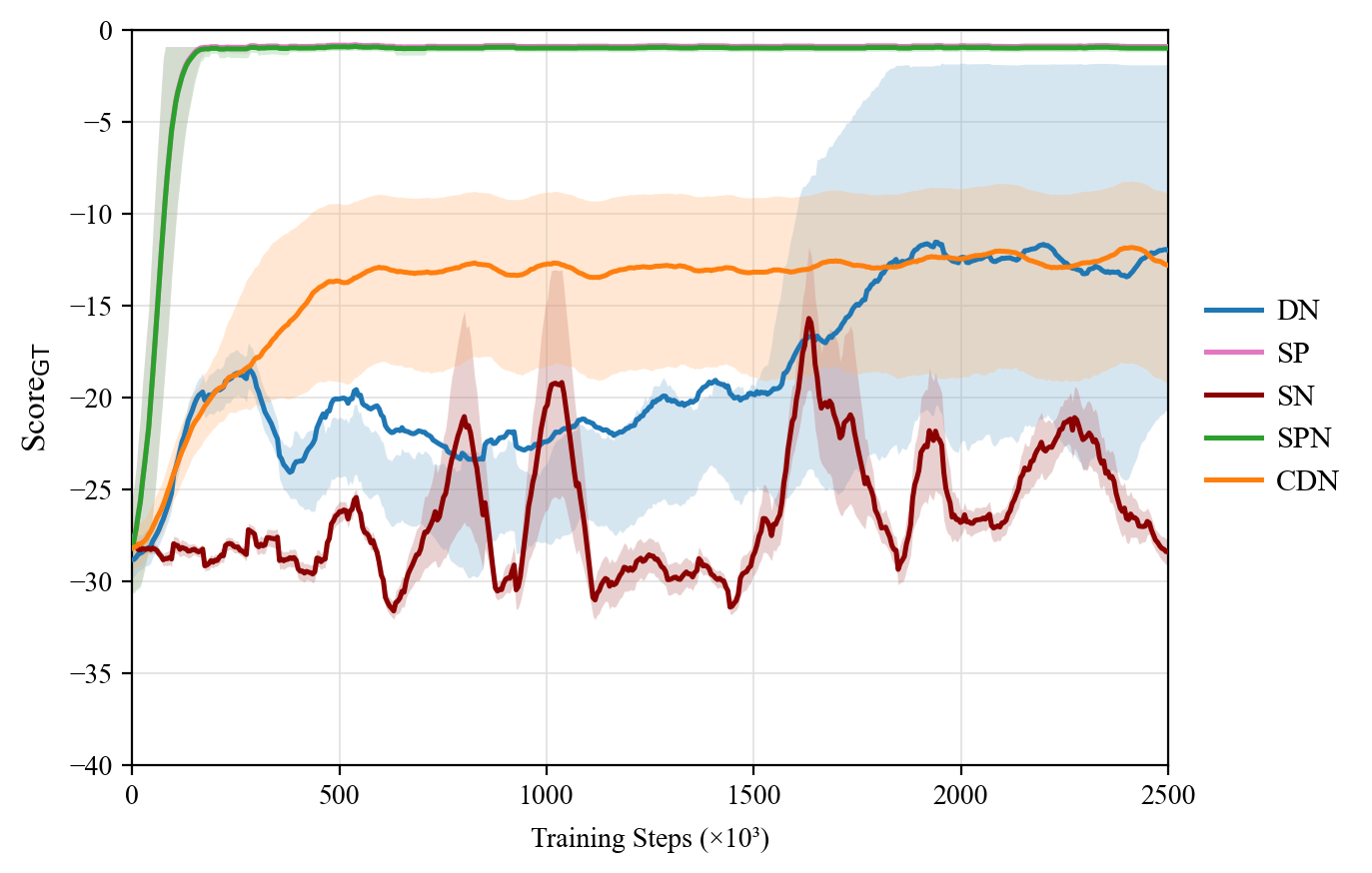} % Replace 'example-image' with your image filename
    \caption{Training curves for the 50 node network size, red then blue agent order in the extended action space for reward functions: sparse positive (SP), sparse negative (SN), sparse positive negative (SPN), dense negative (DN) and complex dense negative (CDN).}
    \label{fig:50_node_training_curve_rb} % Unique label for referencing the figure
\end{figure}

\begin{figure}[!h] % Figure placement: Here, Top, or optional
    \centering % Centers the figure
    \includegraphics[width=0.7\textwidth]{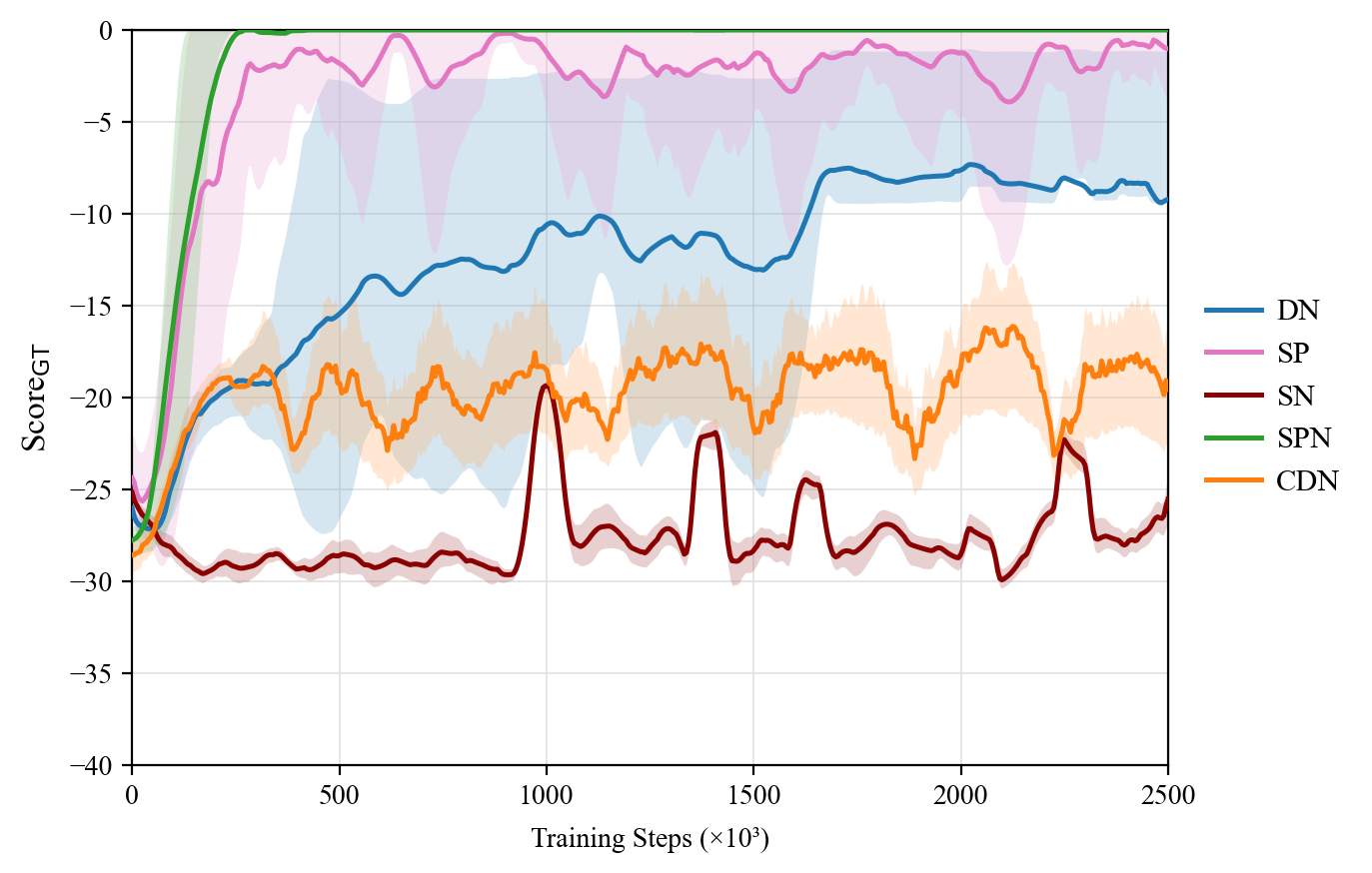} % Replace 'example-image' with your image filename
    \caption{Training curves for the 50 node network size, blue then red agent order in the extended action space for reward functions: sparse positive (SP), sparse negative (SN), sparse positive negative (SPN), dense negative (DN) and complex dense negative (CDN).}
    \label{fig:50_node_training_curve_br} % Unique label for referencing the figure
\end{figure}

\begin{figure}[!h] % Figure placement: Here, Top, or optional
    \centering % Centers the figure
    \includegraphics[width=0.7\textwidth]{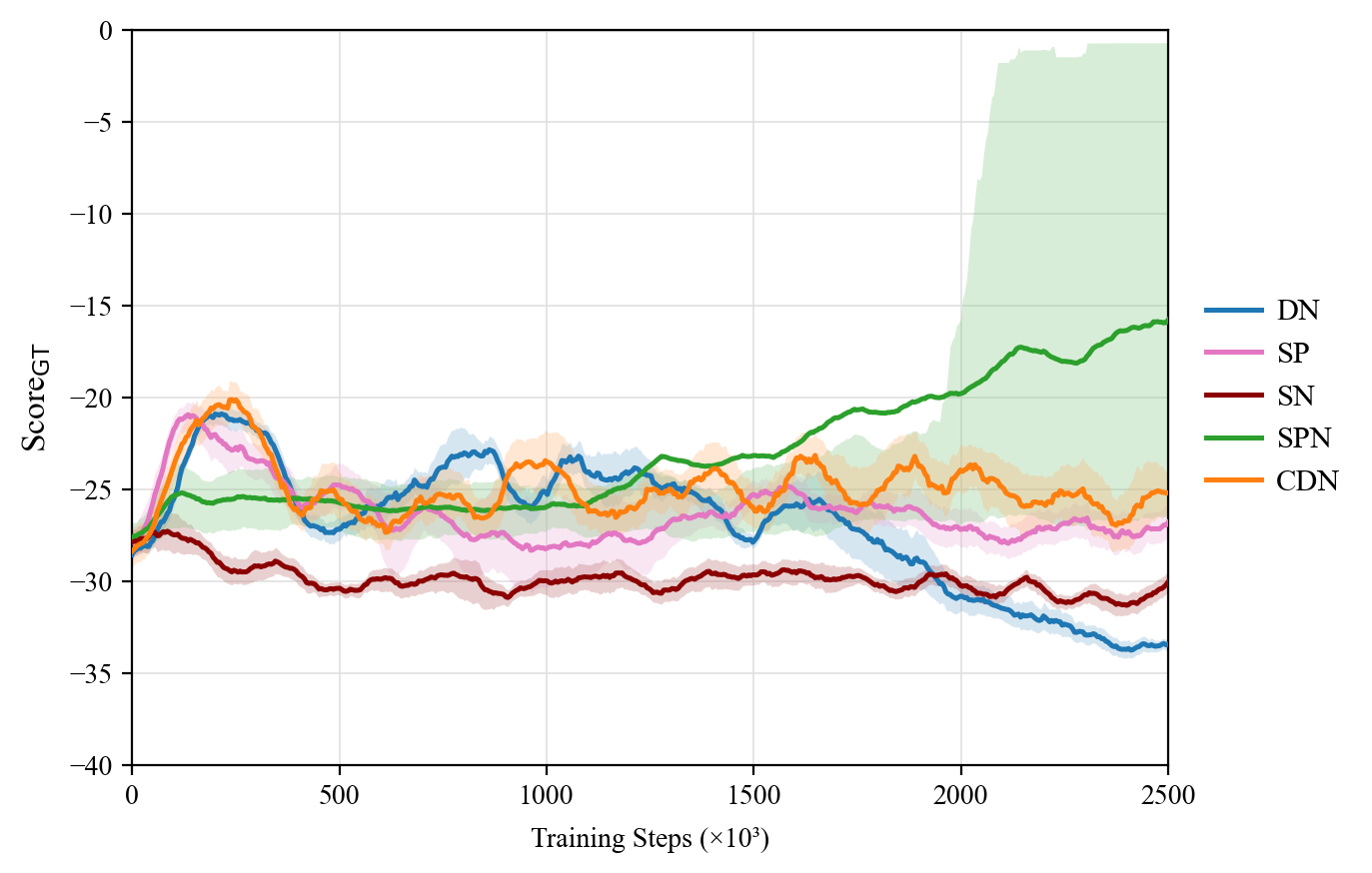} % Replace 'example-image' with your image filename
    \caption{Training curves for the 50 node network size, random agent order in the extended action space for reward functions: sparse positive (SP), sparse negative (SN), sparse positive negative (SPN), dense negative (DN) and complex dense negative (CDN).}
    \label{fig:50_node_training_curve_bal} % Unique label for referencing the figure
\end{figure}

%%%%%%%%%%%%%% Appendix J Temporary miniCAGE results table %%%%%%%%%%%%%

% \clearpage
\section{CAGE agents trained using DQN} \label{miniCAGE_dqn}

Here, in Table~\ref{tab:DQN_General_table_miniCAGE}, we detail our results from evaluating the sparse and dense reward functions in MiniCAGE using DQN.

\begin{table*}[h!]
\centering
\caption{Results for DQN agents trained in MiniCAGE using the sparse positive (SP), sparse negative (SN), sparse positive negative (SPN) and default CAGE reward function (CDN).}
\vspace{0.1in}
\label{tab:DQN_General_table_miniCAGE}
\small
\setlength{\tabcolsep}{4pt}
\begin{tabular}{@{}l c c c c c c c@{}}
\toprule
%\rowcolor{yellow}
\multirow{3}{*}{ \textbf{Reward Function}} 
  & \multirow{3}{*}{\textbf{\begin{tabular}[c]{@{}c@{}}$\text{Score}_\text{GT}$\end{tabular}}} 
  & \multicolumn{4}{c}{\textbf{Average Evaluation Reliability}}  
  & \multicolumn{2}{c}{\textbf{95\% CI}} \\ 
\cmidrule(l){3-6} \cmidrule(l){7-8}
            &              & \textbf{Lower RF} & \textbf{Upper RF} & \textbf{$\bar{\text{DT}}$ (e-3)} & \textbf{$\text{DR}'$} 
            & \textbf{LL} & \textbf{UL} \\ 
\midrule
SP   &1.48	&0.96	&2.97	&0.97	&0.45	&1.42	&1.55 \\
SN   &2.75	&1.78	&3.65	&0.01	&0.28	&2.67	&2.84 \\
 
SPN   &1.51	&0.97	&2.85	&0.97	&0.40	&1.45	&1.58\\
CDN (Default CAGE rewards)    &1.57	&1.02	&2.32	&1.30	&0.43	&1.52	&1.63\\
\bottomrule
\end{tabular}
\end{table*}

%%%%%%%%%%%%%% Appendix K Poitive Reward Ablation Study %%%%%%%%%%%%%

% \clearpage
\section{Numerical Reward Sign Ablation Study} \label{App:PositiveAblation}

To investigate the role of numerical reward sign versus sparsity, we conduct two complementary ablations: (1) an Ablated-SP reward which shifts SP by a constant to remove its positive sign, and (2) a Dense Positive (DP) reward as the counterpart to DN. 

It is notable that rewards which perform poorly in our experiments (CDN, DN, SN) all feature negative penalties without any positive rewards. However, for an idealised RL algorithm the optimal policy is invariant when a constant is added to the reward function~\citep{sutton2018rl, ng99PBRS}.

To empirically determine the role that numerically positive rewards play in the performance of SP and SPN agents (i.e., versus sparsity) we create the Ablated-SP and Dense Positive (DP) reward functions. Ablated-SP simply adds a constant reward (-1) to the SP reward. In other words, the blue agent receives a reward of 0 when the network has zero compromised nodes and -1 otherwise. Similarly, the DP reward is constructed from DN by adding the size of the network so that the reward is +1 per uncompromised node (i.e., 0 when fully compromised only).

%Ablated-SP retains the sparse structure of SP without its positive rewards. DP retains the density of DN whilst replacing its negative penalty with a reward of +1 per uncompromised node. The two ablations together span sparsity and sign, and the results indicate that just the inclusion of positive rewards alone is not the reward trait that leads to better performing agents. 

Using the YT environment, a network size of 10 nodes and all three agent orders, we train each agent for 1.5~million time steps and evaluate the $\text{Score}_\text{GT}$, best $\text{Score}_\text{GT}$ and number of optimal runs.

\begin{table*}[h]
\caption{YT PPO agent results for each agent action order combination in the basic and extended action spaces for network of size 10, comparing agents trained using the Ablated-SP and Dense Positive (DP) reward function with the alternatives.}

% . using SP and the Ablated-SP reward functions.}
\label{tab:Ablation_basic}
\centering
\begin{threeparttable}
\vspace{0.1in}
\small
\setlength{\tabcolsep}{4pt}
\begin{tabular}{@{}p{2.4cm}ccccccccc@{}}
\toprule
\multirow{3}{*}{\rule{0pt}{18mm}\textbf{Reward Function}} & \multicolumn{9}{c}{\textbf{10 Node Network Evaluation}} \\ 
\cmidrule(l){2-10} 
& \multicolumn{3}{c}{\textbf{Red then Blue}} & \multicolumn{3}{c}{\textbf{Blue then Red}} & \multicolumn{3}{c}{\textbf{Random}} \\
\cmidrule(lr){2-4} \cmidrule(lr){5-7} \cmidrule(lr){8-10} 
& \text{\small \begin{tabular}[c]{@{}c@{}}$\text{Score}_\text{GT}$\end{tabular}} 
& \text{\small \begin{tabular}[c]{@{}c@{}}Best \\ $\text{Score}_\text{GT}$\end{tabular}} 
& \text{\small \begin{tabular}[c]{@{}c@{}}No. of \\ Optimal \\ Runs (/25)\end{tabular}} 
& \text{\small \begin{tabular}[c]{@{}c@{}}$\text{Score}_\text{GT}$\end{tabular}} 
& \text{\small \begin{tabular}[c]{@{}c@{}}Best \\ $\text{Score}_\text{GT}$\end{tabular}} 
& \text{\small \begin{tabular}[c]{@{}c@{}}No. of \\ Optimal \\ Runs (/25)\end{tabular}} 
& \text{\small \begin{tabular}[c]{@{}c@{}}$\text{Score}_\text{GT}$\end{tabular}} 
& \text{\small \begin{tabular}[c]{@{}c@{}}Best \\ $\text{Score}_\text{GT}$\end{tabular}} 
& \text{\small \begin{tabular}[c]{@{}c@{}}No. of \\ Optimal \\ Runs (/25)\end{tabular}} \\
\midrule
\multicolumn{10}{@{}l}{\textbf{Basic Action Space}} \\
\midrule
SP         & 0.90&	0.90&	25&	3.91&	0.90&	10&	1.35&	1.34 & \tnote{*} \\
Ablated-SP &0.90  & 0.90&	25&	2.61&  0.90&    20& 1.35&   1.34 & \tnote{*} \\ 
SN         & 6.15 &	0.90&	3&	6.10 &	0.90&	2&	8.35 &  7.62 & \tnote{*} \\
SPN        & 0.90 &	0.90&	25&	0.90&	0.90&	25&	1.35&	1.34 & \tnote{*} \\
DN         & 1.57&	0.90&	12&	1.76&	0.90&	15&	6.85&	1.34 & \tnote{*}\\
CDN        & 2.38&	0.90&	5&	1.53&	0.90&	12&	8.79&	8.54 & \tnote{*} \\
% \rowcolor{yellow!40} 
DP         & 1.49 &	0.90 &	15 & 1.25	&	0.90 &	17 &6.43 & 1.34	& \tnote{*} \\
 
\midrule
\multicolumn{10}{@{}l}{\textbf{Extended Action Space}} \\
\midrule
 
SP          &0.90	&0.90	&25 	&0	    &0	&25	  &0.99&     0.88&\tnote{*}\\
Ablated-SP&  0.90&	0.90&	25&	    0&	    0&	25&	  1.22&	     1.18& \tnote{*}\\
SN &        7.39&	2.87&	0&      6.84&	0&	1&	  8.18&	     6.73& \tnote{*}\\
SPN &        0.90&	0.90&	25&	    0.6&	0&	18&	  1.04&	     0.88& \tnote{*}\\
DN &        1.82&	0.90&	7&	    1.48&	0&	4&	  6.33&	     1.34& \tnote{*}\\
CDN &        2.71&	0.90&	5&	    2.39&	0&	3&	  6.98&	     1.35& \tnote{*}\\
% \rowcolor{yellow!40} 
DP         & 1.64&	0.90&	12&     1.15&	0&	15&   6.80&      1.35& \tnote{*}\\

\bottomrule
\end{tabular}
\begin{tablenotes}
\footnotesize
\item[*]{The optimal policy score is non-trivial so we do not count the number of optimal runs}
\end{tablenotes}
\vspace{-0.5ex}
\end{threeparttable}
\end{table*}

Table~\ref{tab:Ablation_basic} shows that Ablated-SP matches or marginally exceeds SP under the two fixed agent orderings. When the agent order is random, Ablated-SP outperforms the CDN, DN, and SPN rewards, and closely approaches the performance of SP and SPN. Similarly, DP performs comparably to the other dense reward functions across all agent orders. In the Red then Blue order, DP achieves a $\text{Score}_\text{GT}$ of 1.49 in the basic action space and 1.64 in the extended action space, in line with DN (1.57 and 1.82), and substantially worse than SP and SPN (both 0.90) and this relative ranking is repeated in the balanced ordering too.
Taken together, these results show that a numerically positive sign is not the reason why SP and SPN outperform SN, DN, and CDN.

\clearpage
\section{Reward Constant Shift Study} 
\label{App:constant_shift_ablation}

Here we extend the ideas introduced in Appendix~\ref{App:PositiveAblation} and explore a wider range of constant shift factors applied to the SP reward. Whilst the optimal policy is invariant under constant shift~\citep{sutton2018rl, ng99PBRS}, this does not extend to learning dynamics which are highly dependent on the numerical sign, magnitude, algorithm and hyperparameters used during training. 

To explore this further, we characterise the SP reward family $R(\text{safe}){=}1{-}c$, $R(\text{compromised}){=}{-}c$ and then evaluate for $c\in\{-0.5, 0, 0.5, 1.0, 1.5\}$ using YT, a network size of 10, and PPO. 

Initially, we found that $c$ had a significant impact on agent performance and were motivated to understand why. In the standard baselines implementation of PPO, the value and policy updates share a clipping budget. When $c \neq 0$, the value function target becomes $-\dfrac{c}{1-\gamma}$ and causes large value losses, thus clipping both the value and policy weight updates and rendering the learning highly ineffective. We modified PPO to use independent value and policy networks with independent clipping and performed some additional hyperparameter fine tuning. As shown in Table~\ref{tab:Shifted_Rewards}, the results confirm that reward sign is not the main factor determining the performance of trained ACD agents.

%To validate this empirically we modified PPO to clip the value and policy gradient separately. We investigated how a set of constant shifts to the SP reward function affected the $\text{Score}_\text{GT}$, seen in Table~\ref{tab:Shifted_Rewards}. As expected, and in keeping with the conclusion of Appendix K that reward sign is not responsible for the success of a policy, simply shifting the SP reward by a constant c did not result in different $\text{Scores}_\text{GT}$. The only caveat to this can be seen in the balanced agent order since \textcolor{red}{...tbc}. 

% . The reward function is defined as \textbf{$\text{Reward} =  \text{SP} \boldsymbol{-} \boldsymbol{c}$}, where the shift $c$ takes the values shown in the first column
% This appendix is to address the set of reward shifting experiments done (beyond the positive ablation) during the ICML rebuttal. We need a table of results and the table comparing the shifting reward function traits with one another. This positions why SP is actually uniquely good as a reward function.

% When training agents for the Appendix~\ref{App:PositiveAblation}, we found that the SP might provide a uniquely good reward function structure to learn the optimal defensive policy using the PPO algorithm. 

%When we use DRL, incorporating neural networks and using policy gradient, actor-critic algorithms like PPO, the learning dynamics are effected by the techniques that constitute the algorithm, as well as hyperparameters. 

\begin{table*}[h!]
  \caption{Comparing PPO agents trained using shifted SP family reward functions in YT, with a network size of 10.}
 
  \label{tab:Shifted_Rewards}
  \centering
  \begin{threeparttable}
  \vspace{0.1in}
  \small
  \setlength{\tabcolsep}{4pt}
  \begin{tabular}{@{}p{2.4cm}ccccccccc@{}}
  \toprule
  \multirow{3}{*}{\rule{0pt}{18mm}\textbf{Shift $\boldsymbol{c}$}} & \multicolumn{9}{c}{\textbf{10 Node Network Evaluation}} \\
 
  \cmidrule(l){2-10}
  & \multicolumn{3}{c}{\textbf{Red then Blue}} & \multicolumn{3}{c}{\textbf{Blue then Red}} & \multicolumn{3}{c}{\textbf{Balanced}} \\
  \cmidrule(lr){2-4} \cmidrule(lr){5-7} \cmidrule(lr){8-10}
  & \text{\small \begin{tabular}[c]{@{}c@{}}$\text{Score}_\text{GT}$\end{tabular}}
  & \text{\small \begin{tabular}[c]{@{}c@{}}Best \\ $\text{Score}_\text{GT}$\end{tabular}}
  & \text{\small \begin{tabular}[c]{@{}c@{}}No. of \\ Optimal \\ Runs (/25)\end{tabular}}
  & \text{\small \begin{tabular}[c]{@{}c@{}}$\text{Score}_\text{GT}$\end{tabular}}
  & \text{\small \begin{tabular}[c]{@{}c@{}}Best \\ $\text{Score}_\text{GT}$\end{tabular}}
  & \text{\small \begin{tabular}[c]{@{}c@{}}No. of \\ Optimal \\ Runs (/25)\end{tabular}}
  & \text{\small \begin{tabular}[c]{@{}c@{}}$\text{Score}_\text{GT}$\end{tabular}}
  & \text{\small \begin{tabular}[c]{@{}c@{}}Best \\ $\text{Score}_\text{GT}$\end{tabular}}
  & \text{\small \begin{tabular}[c]{@{}c@{}}No. of \\ Optimal \\ Runs (/25)\end{tabular}} \\
  \midrule
  \multicolumn{10}{@{}l}{\textbf{Basic Action Space}} \\
  \midrule
  -0.5 & 0.90 & 0.90 & 25 & 0.90 & 0.90 & 25 & 1.35 & 1.34 & 7 \\
  0    & 0.90 & 0.90 & 25 & 0.90 & 0.90 & 25 & 1.35 & 1.34 & 7 \\
  0.5  & 0.90 & 0.90 & 25 & 0.90 & 0.90 & 25 & 1.35 & 1.34 & 7 \\
  1    & 0.90 & 0.90 & 25 & 0.90 & 0.90 & 25 & 1.35 & 1.34 & 7 \\
  1.5  & 0.90 & 0.90 & 25 & 1.24 & 0.90 & 24 & 1.35 & 1.34 & 7 \\
  \midrule
  \multicolumn{10}{@{}l}{\textbf{Extended Action Space}} \\
  \midrule
  -0.5 & 1.21 & 0.90 & 24 & 0.15 & 0.00 & 23 & 2.02 & 0.88 & 4 \\
  0    & 1.13 & 0.90 & 24 & 0.00 & 0.00 & 25 & 2.69 & 0.88 & 1 \\
  0.5  & 0.90 & 0.90 & 25 & 0.31 & 0.00 & 23 & 2.40 & 0.88 & 3 \\
  1    & 0.94 & 0.90 & 24 & 0.27 & 0.00 & 23 & 2.67 & 0.88 & 2 \\
  1.5  & 1.36 & 0.90 & 23 & 0.00 & 0.00 & 25 & 4.98 & 0.94 & 1 \\
  \bottomrule
  \end{tabular}
  \vspace{-0.5ex}
  \end{threeparttable}
  \end{table*}

% Effective learning dynamics from sparse rewards more generally requires careful consideration of algorithm-specific learning dynamics. 

% =============================================================
% Shift-sensitivity characterisation table (appendix).
% Reward family: R(safe) = 1 - c,  R(compromised) = -c
%   c = 0  <-> SP  (main paper)
%   c = 1  <-> Ablated-SP  (Appendix K)
%
% Requires in preamble:
%   \usepackage{booktabs}
%   \usepackage{multirow}
%   \usepackage{makecell}
% =============================================================
% This should be updated after final evals, knowing what the new gammma value is. Do we need these for both decoy and basic action space? Should we provide a hp table for these experiments?

Additionally, the reason why SP may be a uniquely preferable reward function is shown in Table~\ref{tab:shift_sensitivity}. The value of the worst-case fully-compromised network state is 0 (i.e., closest to its initialisation value) and the clipping ratio is least severe.

\begin{table}[h]
  \centering
  \caption{%
  $V(s_{\text{worst}}){=}{-}c/(1{-}\gamma)$ is the analytical value of the
  worst-case fully-compromised state under a near-absorbing approximation
  at $\gamma{=}0.92$.
  The clipping ratio $\lVert\nabla\mathcal{L}_V\rVert/\lVert\nabla\mathcal{L}_\pi\rVert$
  is the ratio of mean value to mean policy gradient norms, averaged over
  the final 20\% of training across 25 seeds.%
  }
  \vspace{2ex}
  \label{tab:shift_sensitivity}
  \small
  \begin{tabular}{@{}c c c@{}}
  \toprule
  \makecell{\textbf{Constant} \\ \textbf{shift} \\ $\boldsymbol{c}$}                                                                               
  & $\boldsymbol{V(s_{\text{worst}})}$
  & \makecell{\textbf{Clipping} \\ \textbf{ratio}}
    $\left(\dfrac{\lVert\nabla\mathcal{L}_V\rVert}{\lVert\nabla\mathcal{L}_\pi\rVert}\right)$ \\
  \midrule
  $-0.5$           &   6.25  &  7.7 \\
  $0_{\dagger}$    &   0.00  &  4.3 \\
  $0.5$            &  $-6.25$  &  5.7 \\
  $1.0_{\ddagger}$ & $-12.50$  &  9.7 \\
  $1.5$            & $-18.75$  & 14.4 \\
  \bottomrule
  \end{tabular}
  \\[4pt]
  {\footnotesize
  $\dagger$~SP (main paper).\quad
  $\ddagger$~Ablated-SP (Appendix~K).
  }
  \end{table}

%%%%%%%%%%%%%% Appendix M Dense Reward Bias Examples %%%%%%%%%%%%%

\clearpage
\section{Default CAGE-2 Reward Sources of Bias} \label{App:Bias_examples}

%In both the YT and CAGE environmental experiments, the methodology was designed to cover reward structures ranging from extreme sparsity (reward issued only in a single state) through to highly-engineered reward functions, like that of the default CAGE-2 challenge. We believe that these complex reward structures lead to biased blue agents that may not be able to achieve optimal policies due to their lack of alignment with the overarching goal of the ACD task.
The CAGE~2 reward function is dense, highly-engineered and contains potential sources of bias that may lead to misaligned or sub-optimal (e.g., because of a noisy or contradictory reward signals) policies. In contrast to sparse rewards, it is also highly tailored to the specific CAGE-2 challenge scenario and therefore unsuited to modified network configurations without additional work. The specific sources of bias in the CAGE~2 reward function are as follows:

\begin{enumerate}
    \item All compromised user hosts provide the same penalty despite having different vulnerability profiles (and thus different long-term state values). The same is true of hosts in the enterprise subnet. 
    \item The penalty for enterprise hosts and operational server compromise is the same (-1) despite a compromised operational server being much closer to an impacted operational server (-10) from a lateral movement (causal distance from attack objective) perspective.
    \item The penalty for a compromised operational host is -0.1 per time step, the same as a user host, despite requiring fewer steps to reach and impact the operational server. This is an example where the default reward would not generalise well to an adversary that made use of this route, yet a sparse reward would not require modification or domain expertise. 
    \item The cost of performing the restore action is -1, drawing numerical equivalence with the compromise of an enterprise or operational host. It is also equivalent to the compromise of a user host for 10 steps. This means that the resulting policy is biased towards restoring only enterprise or operational hosts (as seen in Table~\ref{tab:blue-action-counts}). This may yield conflicting signals with the fact that compromising user hosts is causally necessary for impacting the operational server, thus failing to restore them leaves the adversary closer to operational impact. Supporting this hypothesis, the SP and SPN rewards use the restore action more sparingly overall and use it mainly on user hosts. 
    % Perhaps mention the issue of arbitrary equivalence (i.e 10 user hosts compromised == 1 enterprise host compromised). 

\end{enumerate}

%%%%%%%%%%%%%% Appendix N - MiniCAGE Agents Evaluation with CAGE-2 Reward Function %%%%%%%%%%%%%
\clearpage

\section{MiniCAGE Evaluation using the Default CAGE-2 Reward} 
\label{App:MiniCAGE_CDN_Eval_Compare}

% TODO: Add one sentence that simply explains What and why is shown. To show... we compare... in Table~\ref{tab:SP_CDN_Comparison}.

In Table~\ref{tab:SP_CDN_Comparison} we show the mean and median scores per timestep using the default CAGE 2 reward across the sets of policies trained using sparse reward functions. The results show that SPN performs similarly to CDN in terms of the mean score (-1.01 vs -0.99), and that SP and SPN perform better in terms of the median (-0.96 vs -1.09). We include GT score for comparison and the results are also supported by the agent policy analysis in Appendix~\ref{App:CAGE_policy_analysis}. 

These results can be understood further by considering Table~\ref{tab:PPO_General_table_miniCAGE} which shows the upper RF of SP and SPN rewards is higher than that of CDN i.e., the worst 5\% of policies have lower scores. This is because there is a higher probability that the operational server is impacted and incurs a large negative penalty. Since operational server impact is causally dependent on user and then enterprise host compromise, and our sparse policies do a much better job of confining adversary impact to the user hosts, we think this may be an exploration issue that could be solved with further hyperparameter tuning. Alternatively, it may be that the optimal way to defend the op server at all costs is by sacrificing enterprise hosts - keeping the adversary ‘stuck’ near the target rather than minimising overall network compromise. This seems untenable for real-world cyber defence.

\begin{table*}[h!]
\centering
\caption{The MiniCAGE agents evaluated over 1000 episodes (one episode is 100 steps) using the $\text{Score}_\text{GT}$ and the original CAGE reward function averaged (Mean and Median) over each timestep.}
\vspace{0.1in}
\label{tab:SP_CDN_Comparison}
\small
\setlength{\tabcolsep}{4pt}
\begin{tabular}{@{}l c c c c@{}}
\toprule
\textbf{Reward Function} 
    & \textbf{$\text{Score}_\text{GT}$} 
 
    & \textbf{\begin{tabular}[c]{@{}c@{}}Mean score per timestep \\ using CAGE~2 default reward\end{tabular}}
    & \textbf{\begin{tabular}[c]{@{}c@{}}Median score per timestep \\ using CAGE~2 default reward\end{tabular}} \\
\midrule
SP   & 1.29 & -1.37 & -0.97 \\
SN   & 2.77 & -2.25 & -2.04 \\
SPN  & 1.35 & -1.01 & -0.96 \\
CDN (CAGE~2 Default) & 1.41 & -0.99 & -1.09\\
 
\bottomrule
\end{tabular}
\end{table*}

\end{document}